\documentclass{article}

\usepackage{microtype}
\usepackage{graphicx}
\usepackage{subcaption}
\usepackage{booktabs} 

\usepackage{hyperref}

\usepackage[preprint]{icml2026}

\usepackage{amsmath}
\usepackage{amssymb}
\usepackage{mathtools}
\usepackage{amsthm}
\usepackage{graphicx}
\usepackage[table,xcdraw]{xcolor}
\usepackage[normalem]{ulem}
\useunder{\uline}{\ul}{}
\usepackage{algorithm}
\usepackage{algorithmic}
\usepackage[capitalize,noabbrev]{cleveref}
\usepackage{enumitem}

\renewcommand{\algorithmicrequire}{\textbf{Input:}}
\renewcommand{\algorithmicensure}{\textbf{Output:}}

\theoremstyle{plain}
\newtheorem{theorem}{Theorem}[section]
\newtheorem{proposition}[theorem]{Proposition}

\theoremstyle{definition}

\theoremstyle{remark}

\usepackage[textsize=tiny]{todonotes}

\icmltitlerunning{Stochastic Voronoi Ensemble for Anomaly Detection}

\begin{document}

\twocolumn[
  \icmltitle{Stochastic Voronoi Ensembles for Anomaly Detection}



  \icmlsetsymbol{equal}{*}

  \begin{icmlauthorlist}
    \icmlauthor{Yang Cao}{yyy,comp}
    \icmlauthor{Sikun Yang}{comp}
    \icmlauthor{Xuyun Zhang}{sch}
    \icmlauthor{Yujiu Yang}{yyy}
  \end{icmlauthorlist}

  \icmlaffiliation{yyy}{Tsinghua Shenzhen International Graduate School, China.}
  \icmlaffiliation{comp}{Great Bay University, China.}
  \icmlaffiliation{sch}{Macquarie University, Australia}

  \icmlcorrespondingauthor{Yang Cao}{charles.cao@ieee.org}

  \icmlkeywords{Machine Learning, ICML}

  \vskip 0.3in
]



\printAffiliationsAndNotice{}  

\begin{abstract}
Anomaly detection aims to identify data instances that deviate significantly from majority of data, which has been widely used in fraud detection, network security, and industrial quality control. Existing methods struggle with datasets exhibiting varying local densities: distance-based methods miss local anomalies, while density-based approaches require careful parameter selection and incur quadratic time complexity. We observe that local anomalies, though indistinguishable under global analysis, become conspicuous when the data space is decomposed into restricted regions and each region is examined independently. Leveraging this geometric insight, we propose SVEAD (Stochastic Voronoi Ensembles Anomaly Detector), which constructs ensemble random Voronoi diagrams and scores points by normalized cell-relative distances weighted by local scale. The proposed method achieves linear time complexity and constant space complexity. Experiments on 45 datasets demonstrate that SVEAD outperforms 12 state-of-the-art approaches.
\end{abstract}

\section{Introduction}

Anomaly detection is the task of identifying instances that deviate significantly from the majority of data, which has been a fundamental problem in machine learning with applications spanning fraud detection, network security, medical diagnosis, and industrial quality control \cite{chandola2009anomaly,liu2024deep,cao2024data,cao2024automation}. The central challenge lies in distinguishing outliers from natural variations when data exhibits heterogeneous structures, multiple clusters, or varying densities.

Existing detection strategies face complementary limitations. Distance-based methods~\citep{angiulli2002fast} efficiently identify global outliers but fail to detect anomalies in regions with varying local densities, as they apply uniform distance thresholds across the entire feature space without adapting to local data concentration. Density-based approaches~\citep{breunig2000lof} address this by measuring local data density to capture local anomalies, but introduce quadratic computational costs and sensitivity to neighborhood parameters. Recent deep learning methods employ autoencoders or generative models to capture complex patterns through reconstruction error or likelihood estimation, yet require substantial training overhead and careful hyperparameter tuning~\cite{pang2021deep}.

\begin{figure}[!htbp]
     \centering
     \begin{subfigure}[b]{0.45\textwidth} 
         \centering
         \includegraphics[width=1\textwidth]{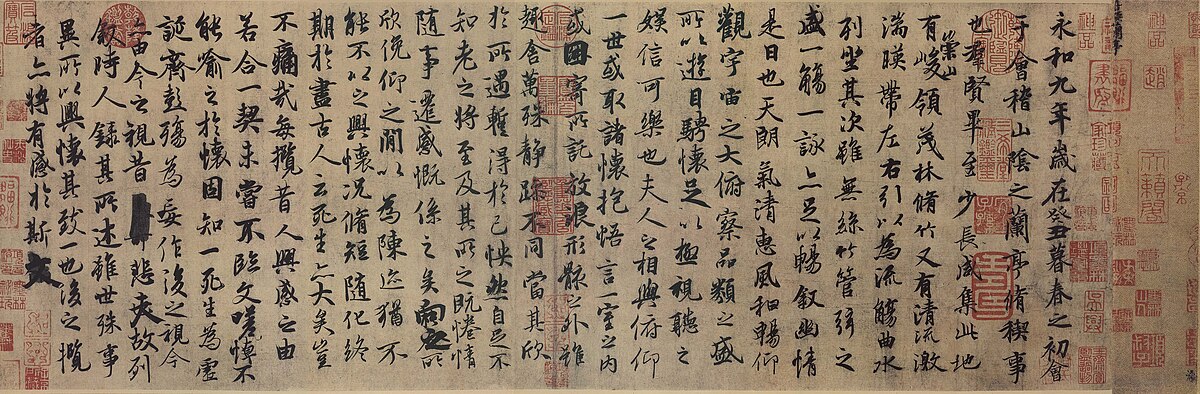} 
         \caption{Original Image.}
     \end{subfigure}

     \begin{subfigure}[b]{0.45\textwidth} 
         \centering
         \includegraphics[width=1\textwidth]{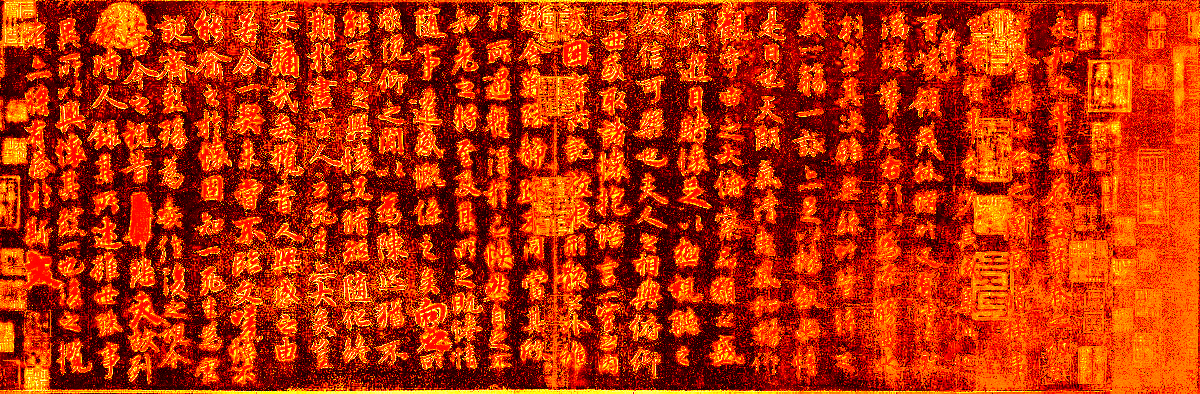} 
         \caption{Isolation Distributional Kernel.}
     \end{subfigure}
     
     \begin{subfigure}[b]{0.46\textwidth}
         \centering
         \includegraphics[width=1\textwidth]{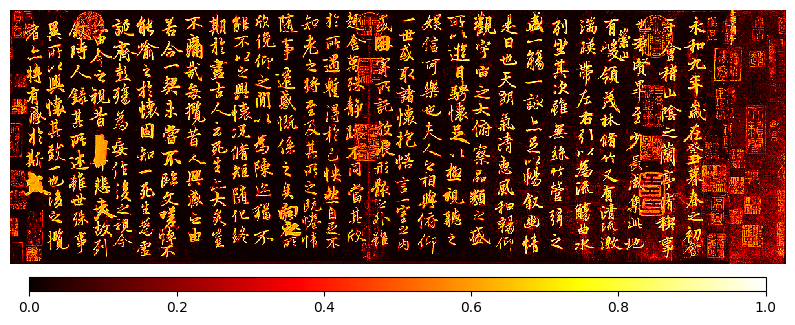}
         \caption{Proposed method: \textbf{SVEAD}.}
     \end{subfigure}
     \caption{Demonstration of anomaly score heatmap on \textit{Lanting Xu} calligraphy pixel data. (a) Original image with black characters considered as anomalies. (b) Anomaly score Heatmap from IDK. (c)Anomaly score Heatmap from SVEAD. Colorbar indicates the anomaly scores.}
    \label{fig:calligraphy}
\end{figure}

These limitations suggest examining the problem from a geometric perspective, which reveals an overlooked property: \textbf{anomalies that appear indistinguishable under global analysis become conspicuous when examined within restricted spatial contexts}. By evaluating samples within dynamically defined spatial regions, we can identify both global and local anomalies through a unified scoring mechanism.

Leveraging this geometric principle, we propose the Stochastic Voronoi Ensembles Anomaly Detector (SVEAD). The proposed algorithm constructs an ensemble of random space partitions via Voronoi diagram. Each partition is generated by randomly sampling a subset of data points as anchors, inducing a Voronoi decomposition where each point is assigned to its nearest anchor. Within each Voronoi cell, we measure how far a point lies from its anchor relative to other points in the same cell. Specifically, distances are normalized by the maximum distance within the cell and weighted by the mean distance, adapting to local scale variations. The final anomaly score is obtained by averaging across all partitions in the ensemble. The method achieves linear time complexity and constant space complexity.

To illustrate detection capability of SVEAD, Figure~\ref{fig:calligraphy} shows results on a section from the classical Chinese calligraphy \textit{Lanting Xu}, where black ink characters are treated as anomalies 
as they deviate from the dominant background distribution and constitute the minority. Compared to the state-of-the-art Isolation Distributional Kernel (IDK)~\cite{ting2020isolation, ting2021isolation}, SVEAD  produces significantly sharper anomaly score maps with clearer character boundaries and better background-foreground separation. 

The choice of Voronoi diagram with random anchor sampling is motivated by three properties: \textbf{1) Data-dependent:} randomly sampling anchors naturally produces more anchors in dense regions and fewer in sparse regions, resulting in smaller cells where data is concentrated and larger cells where data is sparse. \textbf{2) High-efficiency:} Voronoi cells are defined purely by geometric proximity without requiring neighborhood size or distance threshold parameters.
\textbf{3) Non-overlapping:} the diagram divides data space into non-overlapping cells where each point belongs to exactly one cell, ensuring each point receives an unambiguous anomaly score.

The key contributions of this work are:
\begin{itemize}[noitemsep, topsep=0pt]
    \item proposing SVEAD, an ensemble-based algorithm that leverages stochastic Voronoi diagrams to capture multi-scale anomaly patterns without explicit parameter learning or complex model architectures.
    \item introducing a dual-factor scoring mechanism that combines relative position normalization with local density weighting, which enables adaptive anomaly detection across varying density regions without explicit density estimation.
    \item demonstrating the ability SVEAD to detect different types of anomalies, including global, local and dependency anomaly.
    \item conducting extensive experiments on 45 datasets showing that SVEAD achieves state-of-the-art performance while maintaining computational efficiency and interpretability.
\end{itemize}

All the code are available at \url{https://anonymous.4open.science/r/SVEAD-2B97/}.

\section{Problem Formulation}

\textbf{Anomaly Detection Task.} Consider a dataset $\mathcal{X} = \{\mathbf{x}_1, \mathbf{x}_2, \ldots, \mathbf{x}_n\}$ where each instance $\mathbf{x}_i \in \mathbb{R}^d$ is represented by $d$ features. The objective of anomaly detection is to learn a scoring function $f: \mathbb{R}^d \rightarrow \mathbb{R}$ that assigns higher scores to anomalous instances and lower scores to normal instances. 

\textbf{Anomaly Types.} Anomalies are normally categorized into three types\cite{han2022adbench}:

\textit{Global anomalies} $\mathbf{x}_g$ that are distant from all data points. Let $\rho(\mathbf{x})$ denote the local density at point $\mathbf{x}$. A global anomaly satisfies $\rho(\mathbf{x}_g) \ll \mathbb{E}_{\mathbf{x} \in \mathcal{X}}[\rho(\mathbf{x})]$, residing in sparse regions of the feature space with consistently low density across all scales.

\textit{Local anomalies} $\mathbf{x}_l$ that deviate within their neighborhood $\mathcal{N}(\mathbf{x}_l)$ despite appearing normal globally. Formally, $\rho(\mathbf{x}_l) < \mathbb{E}_{\mathbf{x} \in \mathcal{N}(\mathbf{x}_l)}[\rho(\mathbf{x})]$ while $\rho(\mathbf{x}_l) \approx \mathbb{E}_{\mathbf{x} \in \mathcal{X}}[\rho(\mathbf{x})]$. These anomalies exhibit unusual patterns relative to their local context but conform to global statistics.


\textit{Dependency anomalies} $\mathbf{x}_c$ exhibit feature correlation  patterns that deviate from the normal dependency structure. While normal instances satisfy certain feature dependencies characterized by $p(\mathbf{x}) = p(\mathbf{x} | \textit{normal dependencies})$, dependency anomalies violate these correlations. 

\textbf{Unsupervised Setting.} We focus on the unsupervised anomaly detection setting where $\mathcal{X}$ contains an unknown mixture of normal instances (majority class) and anomalous instances (minority class). Let $\mathcal{X} = \mathcal{X}_{\text{normal}} \cup \mathcal{X}_{\text{anom}}$ where $|\mathcal{X}_{\text{anom}}| < |\mathcal{X}_{\text{normal}}|$. The detector must identify $\mathcal{X}_{\text{anom}}$ without access to labeled data, relying solely on the assumption that anomalies deviate from the majority distribution. This contrasts with the semi-supervised one-class setting where $\mathcal{X} = \mathcal{X}_{\text{normal}}$ during training, providing a clean reference of normal behavior.

\section{Related Work}


\subsection{Shallow methods}
Distance-based methods like k-NN~\cite{ramaswamy2000efficient} compute anomaly scores based on distances to nearest neighbors, but fail in datasets with varying densities, where normal points in sparse regions are misclassified as anomalies due to large neighbor distances. 
Density-based approaches address this limitation by examining local density variations. LOF~\cite{breunig2000lof} introduced the concept of local reachability density, comparing each point's density to its neighbors' densities through density ratios. Points with significantly lower local density are identified as anomalies. While LOF handles varying densities better than distance-based methods, it suffers from quadratic time complexity due to k-nearest neighbor computations and requires careful selection of the neighborhood parameter $k$.

Statistical methods take a different approach by modeling feature distributions. COPOD~\cite{li2020copod} uses empirical copula-based modeling to capture feature dependencies while computing tail probabilities for anomaly scoring. ECOD~\cite{li2022ecod} simplifies this by using empirical cumulative distribution functions with feature independence assumptions, estimating tail probabilities for each dimension and aggregating them into anomaly scores. These methods identify points in low-probability regions as anomalies.

Isolation-based methods fundamentally differ from density profiling approaches. Instead of characterizing normal behavior, they exploit the principle that anomalies are easier to isolate. Isolation Forest~\cite{liu2008isolation} constructs an ensemble of random trees that partition data through recursive axis-aligned splits. Anomalies require fewer partitions to isolate, resulting in shorter path lengths. The method provides linear time complexity and parameter-free operation but struggles with local anomalies in dense regions due to its axis-aligned partitioning bias. Recent isolation variants address local anomaly detection. iNNE~\cite{bandaragoda2018isolation} replaces axis-aligned splits with hypersphere-based partitioning, where each hypersphere is centered at a randomly sampled point with radius determined by the nearest neighbor distance. 
IDK~\cite{ting2020isolation,ting2021isolation} further advances this direction by employing data-dependent kernels and kernel mean embeddings to measure distributional similarities~\cite{cao2025anomaly}, which has also been used for time series~\cite{ting2024new,cao2024detecting}, streaming~\cite{cao2024revisiting,xu2025idk} and text~\cite{cao-etal-2025-text} anomaly detection.

\subsection{Deep learning methods}
Deep learning methods leverage neural networks to learn complex data representations for anomaly detection. Autoencoder-based approaches~\cite{aggarwal2016introduction} learn to compress and reconstruct normal data, detecting anomalies through reconstruction error. The underlying assumption is that normal patterns lie on a low-dimensional manifold that the autoencoder learns, while anomalies deviate from this manifold and incur higher reconstruction loss. 

One-class deep learning methods learn compact representations of normal data. DeepSVDD~\cite{ruff2018deep} trains neural networks to map normal instances to a hypersphere with minimal volume in the learned feature space. Anomalies are identified as points falling outside or far from this hypersphere. Graph-based deep methods model relational structures in data. LUNAR~\cite{goodge2022lunar} employs graph neural networks to capture local neighborhood information, learning node representations that encode structural properties for anomaly detection. 

Recent approaches combine deep learning with traditional techniques or introduce novel self-supervised paradigms. DIF~\cite{xu2023deep} uses neural networks for feature extraction and applies isolation forests on learned representations. SLAD~\cite{xu2023fascinating} introduces scale learning for tabular data, randomly sampling feature subspaces and learning to rank representations transformed from varied subspaces through distribution alignment. The method learns inherent data regularities without requiring explicit k-NN computations or density estimation. DTE~\cite{livernochediffusion} estimates the posterior distribution over diffusion timesteps for input samples to detects anomalies.

\section{Methodology}

\begin{table}[!ht]
\centering
\caption{Summary of notation}
\label{tab:notation}
\resizebox{0.48\textwidth}{!}{
\begin{tabular}{cl}
\hline
Symbol & Description \\
\hline
$\mathcal{X}$ & Dataset containing $n$ data points \\
$\mathbf{x}_i$ & The $i$-th data point, $\mathbf{x}_i \in \mathbb{R}^d$ \\
$n$ & Number of data points in dataset \\
$d$ & Feature dimensionality \\
$m$ & Number of anchors per partition \\
$t$ & Ensemble size (number of partitions) \\
$\mathcal{A}^{(k)}$ & Anchor set for the $k$-th partition \\
$a_i^{(k)}$ & The $i$-th anchor in the $k$-th partition \\
$C_i^{(k)}$ & Voronoi cell of anchor $a_i$ in the $k$-th partition \\
$\mathcal{P}^{(k)}$ & The $k$-th partition, $\mathcal{P}^{(k)} = \{C_1^{(k)}, \ldots, C_m^{(k)}\}$ \\
$\delta$ & Distance from point to its assigned anchor \\
$\delta_{\max}$ & Maximum distance from anchor to points within cell \\
$\delta_{\text{mean}}$ & Mean distance from anchor to points within cell \\
$s^{(k)}(\mathbf{x})$ & Anomaly score of $\mathbf{x}$ in the $k$-th partition \\
$f(\mathbf{x})$ & Final anomaly score of $\mathbf{x}$ (ensemble average) \\
$\rho(\mathbf{x})$ & Local density at point $\mathbf{x}$ \\
$\mathcal{N}(\mathbf{x})$ & Neighborhood of point $\mathbf{x}$ \\
$\tau$ & Anomaly detection threshold \\
\hline
\end{tabular}}
\end{table}

In this section, we introduce SVEAD (Stochastic Voronoi Ensemble Anomaly Detector), a geometric approach to anomaly detection that leverages random space partitioning through Voronoi diagrams. The core insight is that \emph{anomalies reveal themselves through their positions within locally defined spatial regions: points residing far from local anchors, particularly in sparse cells, are more likely to be anomalous.} Table~\ref{tab:notation} summarizes the key notation used throughout this paper.

SVEAD operates in three stages. First, we construct an ensemble of Voronoi diagrams by repeatedly sampling random subsets of data points as anchors. Each anchor set induces a partition where every point is assigned to its nearest anchor, creating a collection of Voronoi cells. Second, within each cell, we score points by combining their normalized distance to the anchor with the cell's mean distance, yielding a measure that adapts to local density variations. Third, we aggregate scores across all partitions through ensemble averaging, producing robust anomaly assessments that capture both global and local deviations.

This design offers three key advantages. \textbf{1)} Random anchor sampling provides computational efficiency, requiring only nearest-anchor queries rather than complete diagram construction. \textbf{2)} Voronoi diagrams supplies density-adaptive partitioning, where cell sizes naturally reflect local data concentration. \textbf{3)} Ensemble averaging delivers robustness, allowing different random partitions to capture complementary local structures while mitigating sensitivity to any single partition.

\begin{figure*}[!htbp]
    \centering
    \includegraphics[width=0.8\textwidth]{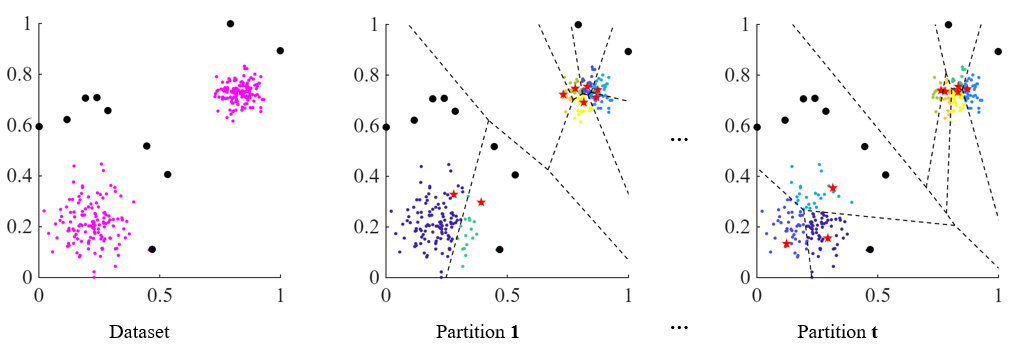}
    \caption{Illustration of SVEAD's space partition and ensemble approach. Left: Original dataset with normal points (purple) and anomalies (black). Middle and Right: demonstration of two random partitions with different anchor sets (red stars). Each partition creates distinct Voronoi cells (dashed lines), assigning points to nearest anchors.}
    \label{fig:step}
\end{figure*}
\subsection{Stochastic Voronoi diagrams via anchor sampling}


We adopt a sampling-based approach that constructs implicit Voronoi diagrams efficiently, Figure~\ref{fig:step} shows the anchor sampling and voronoi diagram ensembles.

\textbf{Anchor Sampling.} For each partition, we randomly sample $m$ points from the dataset $\mathcal{X}$ without replacement to form an anchor set $\mathcal{A} = \{a_1, a_2, \ldots, a_m\}$ where $m \ll n$. These anchors serve as generators for the Voronoi diagrams. The sampling is performed uniformly, ensuring that dense regions naturally receive more anchors due to higher point concentration, while sparse regions receive fewer anchors. This automatic density adaptation occurs without explicit density estimation.

\textbf{Voronoi Cell Assignment.} Given anchor set $\mathcal{A}$, the Voronoi cell $C_i$ associated with anchor $a_i$ is defined as:
\begin{equation}
C_i = \{\mathbf{x} \in \mathcal{X} : \|\mathbf{x} - a_i\| \leq \|\mathbf{x} - a_j\|, \forall j \neq i\}
\end{equation}
where $\|\cdot\|$ denotes the Euclidean distance. Each point $\mathbf{x} \in \mathcal{X}$ is assigned to its nearest anchor, creating a partition $\{C_1, C_2, \ldots, C_m\}$ that satisfies two properties: (1) $\bigcup_{i=1}^m C_i = \mathcal{X}$, ensuring complete coverage, and (2) $C_i \cap C_j = \emptyset$ for $i \neq j$, ensuring exclusivity. Unlike explicit Voronoi diagram construction, we do not compute cell boundaries or vertices; we only determine point-to-cell membership through nearest-anchor queries.

\textbf{Ensemble Construction.} A single random partition may be unstable, as different anchor samples can produce vastly different cell configurations. To achieve robustness, we repeat the sampling and partitioning process $t$ times, generating an ensemble of diagrams $\{\mathcal{P}^{(1)}, \mathcal{P}^{(2)}, \ldots, \mathcal{P}^{(t)}\}$ where each $\mathcal{P}^{(k)}, k\in 1\dots t$ represents the $k$-th partition with its own anchor set $\mathcal{A}^{(k)}$ and cell assignment. Each partition provides an independent view of the data's local structure. Points that consistently appear far from anchors across multiple partitions are more likely to be genuine anomalies, while random fluctuations in individual partitions are averaged out.

\subsection{Anomaly Scoring Mechanism}

The scoring mechanism operates within each Voronoi cell, measuring how anomalous a point appears relative to its local context. A naive approach would directly use the distance to the nearest anchor, but this \textbf{fails to account for varying cell sizes and local density patterns}. We introduce a dual-factor scoring scheme that addresses these challenges.

\textbf{Within-Cell Scoring.} Consider a point $\mathbf{x} \in C_i$ assigned to anchor $a_i$ in cell $C_i$. We define its anomaly score within this partition as:
\begin{equation}
s(\mathbf{x}) = \frac{\|\mathbf{x} - a_i\|}{\max_{\mathbf{x}' \in C_i} \|\mathbf{x}' - a_i\|} \cdot \frac{1}{|C_i|} \sum_{\mathbf{x}' \in C_i} \|\mathbf{x}' - a_i\|
\end{equation}~\label{score}
where the first term $\delta/\delta_{\max}$ normalizes the distance by the maximum cell radius, and the second term $\delta_{\text{mean}}$ represents the mean distance within the cell. We denote $\delta = \|\mathbf{x} - a_i\|$, $\delta_{\max} = \max_{\mathbf{x}' \in C_i} \|\mathbf{x}' - a_i\|$, and $\delta_{\text{mean}} = \frac{1}{|C_i|} \sum_{\mathbf{x}' \in C_i} \|\mathbf{x}' - a_i\|$, yielding the compact form from Equation~\ref{score}:
\begin{equation}
s(\mathbf{x}) = \frac{\delta}{\delta_{\max}} \cdot \delta_{\text{mean}},
\end{equation}
In the rare case where a cell contains only the anchor ($|C_i|=1$), we skip scoring for this point in this partition and rely on the other $t-1$ partitions in the ensemble average.

\textbf{Rationale for Dual Factors.} Each component serves a distinct purpose in anomaly assessment:

\textit{Relative position normalization} ($\delta/\delta_{\max}$): This term measures where a point lies within its cell, producing values in $[0, 1]$. Without normalization, raw distances are incomparable across cells of different sizes. A point 10 units from its anchor in a large sparse cell may be more normal than a point 2 units away in a small dense cell. Normalization by $\delta_{\max}$ eliminates this scale dependency, treating each cell as a unit hypersphere where edge points receive scores near 1 regardless of absolute cell size.

\textit{Local density weighting} ($\delta_{\text{mean}}$): If we used only $\delta/\delta_{\max}$, every cell would have points with scores near 1 (those at cell boundaries), incorrectly treating boundary points in all cells as equally anomalous. The mean distance $\delta_{\text{mean}}$ corrects this by encoding local scale. Due to uniform anchor sampling, dense regions receive more anchors (producing smaller cells with smaller $\delta_{\text{mean}}$), while sparse regions receive fewer anchors (producing larger cells with larger $\delta_{\text{mean}}$). In dense regions where points cluster tightly around anchors, $\delta_{\text{mean}}$ is small, suppressing anomaly scores. In sparse regions where points spread widely, $\delta_{\text{mean}}$ is large, amplifying scores. This weighting ensures that points at cell edges in sparse regions receive higher scores than edge points in dense regions.

\textbf{Connection to Density Estimation.} The mean distance $\delta_{\text{mean}}$ naturally encodes local density information. Consider a Voronoi cell containing $|C_i|$ points distributed around anchor $a_i$. If these points occupy a volume $V_i$ in feature space, the local density is approximately $\rho_i \propto |C_i|/V_i$. Assuming points are roughly uniformly distributed within the cell, the mean distance relates to volume through $V_i \propto \delta_{\text{mean}}^d$ where $d$ is the dimensionality. Therefore:
\begin{equation}
\delta_{\text{mean}} \propto V_i^{1/d} \propto \left(\frac{|C_i|}{\rho_i}\right)^{1/d} \propto \rho_i^{-1/d}
\end{equation}
Thus $\delta_{\text{mean}}$ increases as density decreases, providing an adaptive density estimator without explicit k-NN computations. The complete scoring function $(\delta/\delta_{\max}) \cdot \delta_{\text{mean}}$ combines relative position with density adaptation, assigning high scores to points that are both far from their local anchor and located in sparse regions.

\textbf{Ensemble Aggregation.} Across $t$ partitions, a point $\mathbf{x}$ receives $t$ scores $\{s^{(1)}(\mathbf{x}), s^{(2)}(\mathbf{x}), \ldots, s^{(t)}(\mathbf{x})\}$. The final anomaly score is computed as the arithmetic mean:
\begin{equation}
f(\mathbf{x}) = \frac{1}{t} \sum_{k=1}^t s^{(k)}(\mathbf{x})
\end{equation}
Ensemble aggregation stabilizes scores by reducing variance introduced by random anchor sampling. Genuine anomalies consistently receive high scores across partitions because they remain distant from anchors and reside in sparse cells regardless of the specific anchor configuration. Normal points exhibit more variation, as some partitions may place anchors nearby while others do not, but their averaged scores remain low due to overall proximity to the data distribution.

\subsection{Theoretical Properties}

We establish theoretical properties that explain why SVEAD effectively identifies anomalies through random Voronoi diagrams and ensemble averaging.

\textbf{Density Adaptation Property.} The expected cell size in random Voronoi diagrams naturally reflects local data density, providing automatic scale adaptation without explicit density estimation.

\begin{proposition}[Cell Size and Density]
\label{prop:density}
Consider a region $R \subset \mathbb{R}^d$ with local density $\rho(R)$. When uniformly sampling $m$ anchors from dataset $\mathcal{X}$, the expected number of anchors falling in $R$ is proportional to $\rho(R) \cdot |R|$, where $|R|$ denotes the volume of $R$. Consequently, the expected Voronoi cell size in $R$ is inversely proportional to $\rho(R)$.
\end{proposition}
The following analysis assumes that the number of anchors $m$ is sufficiently large and that boundary effects are negligible.
\begin{proof}
Let $N_R$ denote the number of data points in region $R$, and let $M_R$ denote the number of anchors falling in $R$. Under uniform sampling, the probability that any sampled anchor falls in $R$ is $p_R = N_R/n$. The expected number of anchors in $R$ is:
\begin{equation}
\mathbb{E}[M_R] = m \cdot p_R = m \cdot \frac{N_R}{n}
\end{equation}
Since $N_R \approx \rho(R) \cdot |R|$ for sufficiently large $N_R$, we have:
\begin{equation}
\mathbb{E}[M_R] \propto \rho(R) \cdot |R|
\end{equation}
The expected cell size in $R$ is the volume divided by the number of anchors:
\begin{equation}
\mathbb{E}[\text{cell size in } R] \approx \frac{|R|}{\mathbb{E}[M_R]} \propto \frac{1}{\rho(R)}
\end{equation}


Thus, higher density regions produce smaller cells, and lower density regions produce larger cells.
\end{proof}

This property ensures that $\delta_{\text{mean}}$ serves as a valid density proxy: in dense regions, smaller cells yield smaller $\delta_{\text{mean}}$, while in sparse regions, larger cells yield larger $\delta_{\text{mean}}$.

\textbf{Ensemble Convergence.} Averaging across multiple random partitions reduces variance and improves anomaly detection reliability.

\begin{proposition}[Score Variance Reduction]
\label{prop:variance}

Let $s^{(k)}(\mathbf{x})$ denote the anomaly score of point $\mathbf{x}$ in the $k$-th partition with variance $\sigma^2$. While scores across partitions exhibit some correlation due to anchors being sampled from the same dataset, treating them as approximately independent for analytical purposes yields:
\begin{equation}
\text{Var}[f(\mathbf{x})] \approx \frac{\sigma^2}{t}.
\end{equation}
Thus, increasing ensemble size $t$ reduces score variance by a factor of $1/t$.
\end{proposition}

This variance reduction ensures that genuine anomalies, which consistently receive high scores across partitions, are reliably distinguished from normal points whose scores fluctuate randomly.


\textbf{Separation Between Anomalies and Normal Points.} We analyze the expected score difference between anomalies and normal points.

\begin{proposition}[Expected Score Separation]
\label{prop:separation}
Consider a global anomaly $\mathbf{x}_g$ residing in a sparse region with density $\rho_{\text{sparse}}$ and a normal point $\mathbf{x}_{nor}$ in a dense region with density $\rho_{\text{dense}}$ where $\rho_{\text{sparse}} \ll \rho_{\text{dense}}$. 
Anomalies tend to:
1) Reside in sparse regions where cells are large;
2) Lie at cell boundaries, far from anchors.
The expected cell mean distances satisfy:
\begin{equation}
\mathbb{E}[\delta_{\text{mean}}(\mathbf{x}_g)] \propto \rho_{\text{sparse}}^{-1/d} \gg \rho_{\text{dense}}^{-1/d} \propto \mathbb{E}[\delta_{\text{mean}}(\mathbf{x}_{nor})]
\end{equation}
where $d$ is the feature dimensionality. Combined with the relative position term, anomalies receive significantly higher scores.
\end{proposition}

This proposition formalizes the intuition that sparse regions amplify anomaly scores through the $\delta_{\text{mean}}$ weighting factor, creating clear separation between anomalous and normal instances.




These theoretical properties collectively explain SVEAD's effectiveness: random sampling provides density adaptation (Proposition~\ref{prop:density}), ensemble averaging reduces variance (Proposition~\ref{prop:variance}) and the dual-factor scoring creates clear separation between anomalies and normal points (Proposition~\ref{prop:separation}).

\subsection{Algorithm and Complexity Analysis}

\begin{algorithm}[!ht]
\caption{SVEAD}
\label{alg:svead}
\begin{algorithmic}[1]
\REQUIRE Dataset $\mathcal{X} = \{\mathbf{x}_1, \ldots, \mathbf{x}_n\}$, anchor count $m$, ensemble size $t$
\ENSURE Anomaly scores $\{f(\mathbf{x}_1), \ldots, f(\mathbf{x}_n)\}$
\FOR{$k = 1$ to $t$}
    \STATE Sample anchor set $\mathcal{A}^{(k)} = \{a_1^{(k)}, \ldots, a_m^{(k)}\}$ uniformly from $\mathcal{X}$
    \FOR{each $\mathbf{x}_i \in \mathcal{X}$}
        \STATE $j^* \leftarrow \arg\min_{j \in [m]} \|\mathbf{x}_i - a_j^{(k)}\|$
        \STATE Assign $\mathbf{x}_i$ to cell $C_{j^*}^{(k)}$
    \ENDFOR
    \FOR{each cell $C_j^{(k)}$}
        \STATE $\delta_{\max}^{(k,j)} \leftarrow \max_{\mathbf{x} \in C_j^{(k)}} \|\mathbf{x} - a_j^{(k)}\|$
        \STATE $\delta_{\text{mean}}^{(k,j)} \leftarrow \frac{1}{|C_j^{(k)}|} \sum_{\mathbf{x} \in C_j^{(k)}} \|\mathbf{x} - a_j^{(k)}\|$
    \ENDFOR
\ENDFOR
\FOR{each $\mathbf{x}_i \in \mathcal{X}$}
    \STATE Initialize $f(\mathbf{x}_i) \leftarrow 0$
    \FOR{$k = 1$ to $t$}
        \STATE Let $j^*$ be the cell index where $\mathbf{x}_i \in C_{j^*}^{(k)}$
        \STATE $\delta \leftarrow \|\mathbf{x}_i - a_{j^*}^{(k)}\|$
        \STATE $s^{(k)}(\mathbf{x}_i) \leftarrow \frac{\delta}{\delta_{\max}^{(k,j^*)}} \cdot \delta_{\text{mean}}^{(k,j^*)}$
        \STATE $f(\mathbf{x}_i) \leftarrow f(\mathbf{x}_i) + s^{(k)}(\mathbf{x}_i)$
    \ENDFOR
    \STATE $f(\mathbf{x}_i) \leftarrow f(\mathbf{x}_i) / t$
\ENDFOR
\ENSURE $\{f(\mathbf{x}_1), \ldots, f(\mathbf{x}_n)\}$
\end{algorithmic}
\end{algorithm}

\textbf{Algorithm Description.} Algorithm~\ref{alg:svead} presents the complete procedure of SVEAD. The algorithm consists of two phases: ensemble construction and anomaly scoring. In the construction phase (lines 1-10), we generate $t$ random partitions. For each partition, we sample $m$ anchors uniformly from the dataset and assign each point to its nearest anchor, computing cell statistics including maximum and mean distances. In the scoring phase (lines 11-18), we iterate through all points and partitions, computing within-cell scores using the dual-factor formula and averaging across partitions to obtain final anomaly scores.

\textbf{Time Complexity.} We analyze the computational cost for each algorithmic component:

\textit{Partition construction} (lines 1-10): Sampling $m$ anchors requires $O(m)$ time. For each of $n$ points, finding the nearest anchor among $m$ candidates requires $O(md)$ operations where $d$ is the feature dimensionality, taking $O(nmd)$ in total. Constructing one partition thus requires $O(nmd)$, and building $t$ partitions costs $O(nmdt)$.

\textit{Score computation} (lines 11-19): For each point and each partition, we recompute the cell assignment by finding the nearest anchor, requiring O(md) operations. Computing the score requires O(d). Processing all points across all partitions requires O(nmdt) operations.

The total time complexity is $O(nmdt) + O(nmdt) = O(nmdt)$, which simplifies to $O(mnt)$ when $d$ is treated as a constant or is small relative to $n$. Since $m \ll n$ in practice, this scales \textbf{linearly} with dataset size.

\textbf{Space Complexity.} The algorithm stores anchor sets across $t$ partitions requiring $O(tmd)$ space, and cell statistics ($\delta_{\max}$ and $\delta_{\text{mean}}$) for $m$ cells per partition requiring $O(tm)$ space. The total space complexity is $O(tmd + tm) = O(tmd)$, which simplifies to $O(tm)$ when $d$ is constant. Excluding the input dataset itself, the algorithm's memory footprint is $O(tm)$, where $t$ and $m$ are constant hyperparameters, thus SVEAD's space complexity is \textbf{constant}.

\section{Experiments}
This section conducts a comprehensive evaluation with the existing state-of-the-art methods to empirically verify the effectiveness and efficiency of SVEAD. Our evaluation covers a wide range of datasets, including synthetic datasets with different anomaly types and 45 real-world datasets.

\subsection{Experimental Setup}

\textbf{Datasets.} We evaluate SVEAD on 45 publicly available benchmark datasets widely used in anomaly detection from ADBench~\cite{han2022adbench}. These datasets span diverse domains including medical diagnosis, network security, image recognition, and industrial monitoring, with varying characteristics in terms of dimensionality (ranging from 3 to 1,555 features), sample size (from 80 to 619,326 instances), and anomaly ratio (from 0.17\% to 34.99\%). 

\textbf{Baseline Methods.} We compare SVEAD against 12 state-of-the-art anomaly detection methods spanning different paradigms: traditional methods including LOF~\cite{breunig2000lof}, Isolation Forest~\cite{liu2008isolation}, iNNE~\cite{bandaragoda2018isolation}, IDK~\cite{ting2020isolation}, ECOD~\cite{li2022ecod}, and COPOD~\cite{li2020copod}; deep learning methods including DeepSVDD~\cite{ruff2018deep}, Autoencoder (AE), LUNAR~\cite{goodge2022lunar}, DIF~\cite{xu2023deep}, SLAD~\cite{xu2023fascinating}, and DTE~\cite{livernochediffusion}. Most algorithms are implemented from the PyOD library \cite{zhao2019pyod}, while SLAD and DTE are obtained from their official repositories.

\textbf{Evaluation Metrics.} We adopt Area Under the Receiver Operating Characteristic curve (AUC-ROC) and Area Under the Precision-Recall curve (AUC-PR) in the experiment. For each dataset and method, we repeat experiments 5 times and report the average performance. Following standard unsupervised anomaly detection evaluation protocol, the labels are used to compute the AUC-ROC and AUC-PR for each dataset only after the models have made predictions. This protocol is consistent with established benchmarking practices in unsupervised anomaly detection 
\citep{han2022adbench, zhao2019pyod}.

\textbf{Parameter Settings.} For fair comparison, we conduct grid search for methods with tunable parameters. LOF's neighborhood size $k$ is searched over $\{5, 10, 20, 40\}$. For Isolation Forest, iNNE, IDK, and SVEAD, the number of sample points is searched over $\{2^1, 2^2, \ldots, 2^8\}$, with ensemble size fixed at 100. ECOD and COPOD are parameter-free and use default settings. Deep learning methods employ default hyperparameters reported in their original papers due to their extensive parameter spaces. 

\subsection{Overall Performance and Statistical Analysis}

\begin{figure}[!ht]
     \centering
     \begin{subfigure}[b]{0.23\textwidth}
         \centering
         \includegraphics[width=1\textwidth]{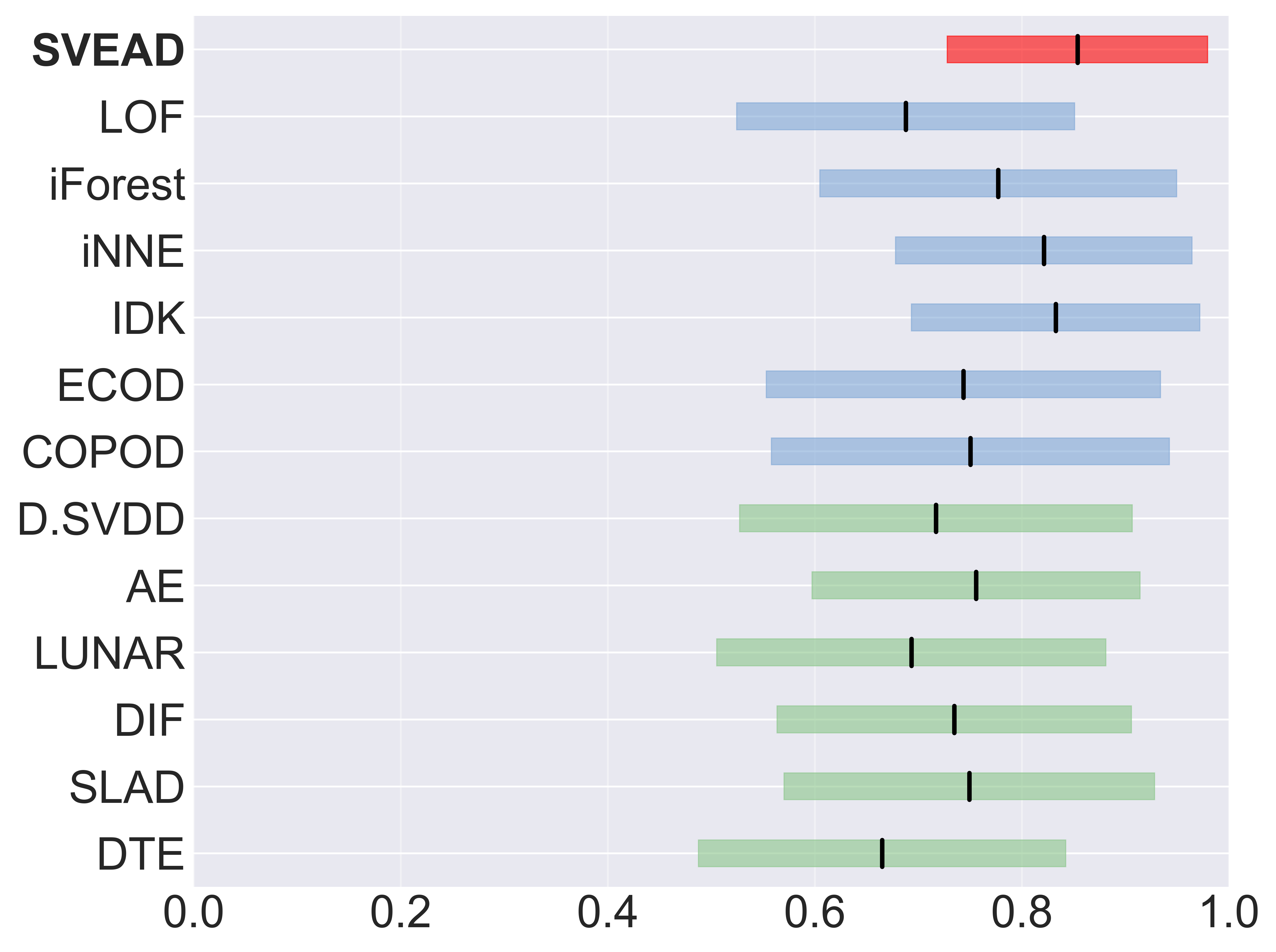}
         \caption{AUC-ROC}
     \end{subfigure}
     \begin{subfigure}[b]{0.23\textwidth}
         \centering
         \includegraphics[width=1\textwidth]{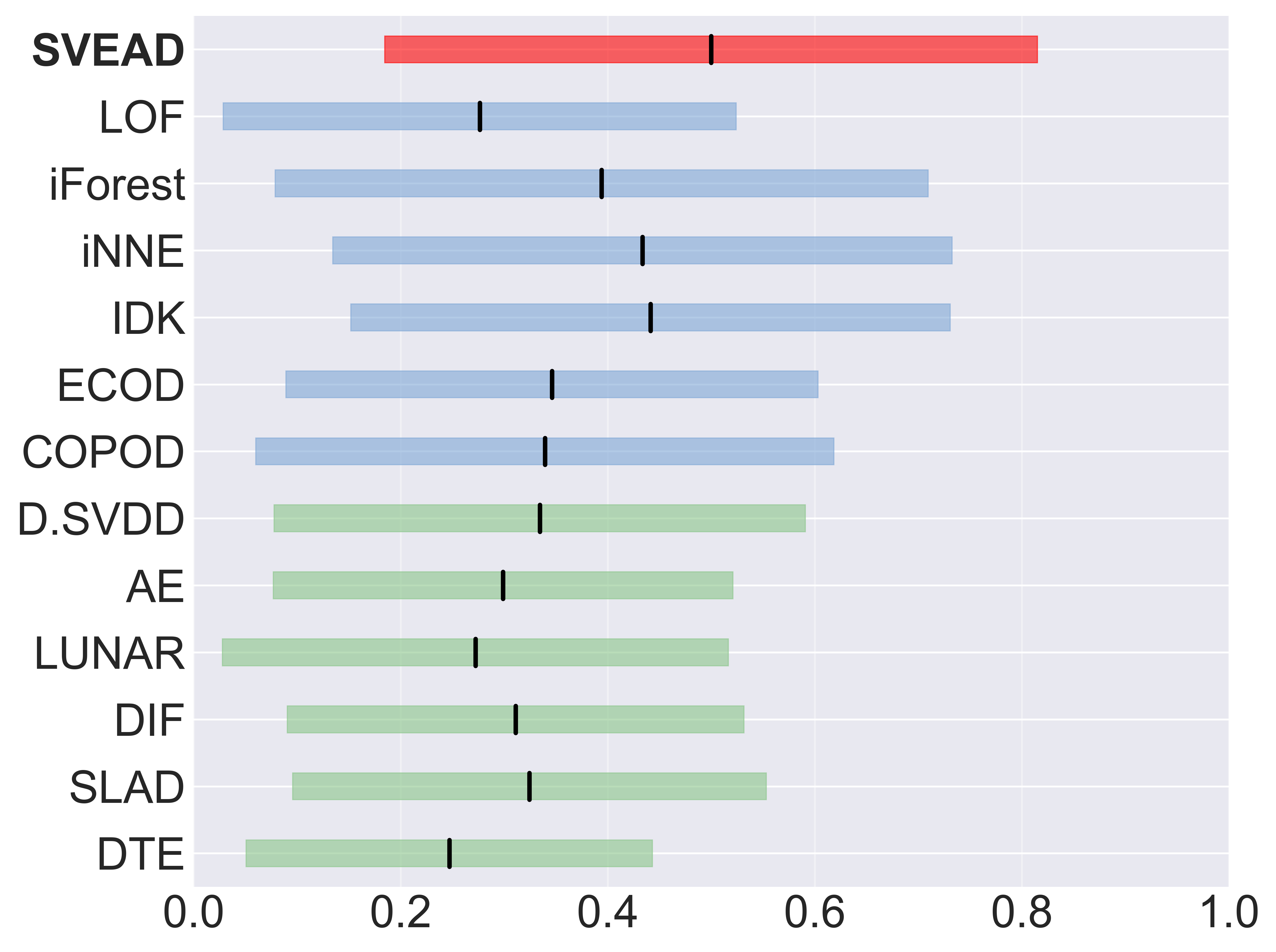}
         \caption{AUC-PR}
     \end{subfigure}
     \caption{Performance comparison across 45 benchmark datasets. (a) AUC-ROC and (b) AUC-PR showing mean and standard deviation over 5 runs. The black vertical line indicates the mean value. SVEAD (red) outperforms shallow methods (blue) and deep learning methods (green) on both metrics.}
     \label{fig:overall}
\end{figure}

Figure~\ref{fig:overall} compares the overall performance across 45 datasets. SVEAD achieves the highest average performance on both AUC-ROC and AUC-PR, outperforming all baselines. Among classical methods, isolation-based approaches (iNNE, IDK) show competitive performance. Deep learning methods achieve lower performance because most of them (e.g. autoencoder) assume clean training data to learn normal patterns. In the unsupervised setting with contaminated samples, the learned representations become unreliable, resulting in poor detection accuracy. In contrast, SVEAD directly scores points based on local relative positions and density without requiring any learning process.

\textbf{Performance on Different Anomaly Types.} To assess SVEAD's ability to detect diverse anomaly patterns, we construct three synthetic datasets representing canonical anomaly types: global, local and dependency anomalies. Table~\ref{tab:anomaly_types} visualizes detection results using contour plots 
of anomaly scores. The results demonstrate SVEAD's versatility across anomaly types. On global anomalies, SVEAD, IDK, and Isolation Forest all successfully identify anomalies in sparse regions, confirming that geometric methods handle globally isolated points effectively. On local anomalies, SVEAD produces significantly sharper boundaries around local deviations compared to IDK, while Isolation Forest struggles due to its axis-aligned partitioning bias. On dependency anomalies, SVEAD captures the anomalous structure more accurately than both baselines. 

\begin{table}[!htbp]
    \centering
    \caption{Performance comparison of iForest and IDK on global (S1), local (S2) and dependency anomaly (S3) with contour plot. Normal and anomounous points are in white and black, and yellow is normal region indicated by each method.}
    \label{tab:anomaly_types}
    \resizebox{0.45\textwidth}{!}{
    \begin{tabular}{cccc}
    \hline 
          &   \textbf{iForest}   &   \textbf{IDK} & \textbf{SVEAD} \\
    \hline 
        \textbf{S1}  & \includegraphics[width=0.2\textwidth]{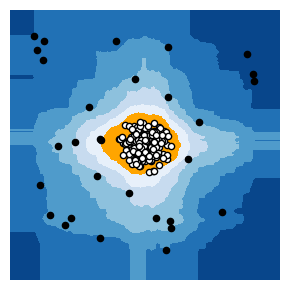} & \includegraphics[width=0.2\textwidth]{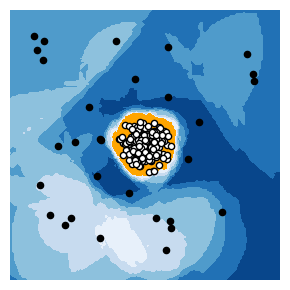}  &
        \includegraphics[width=0.2\textwidth]{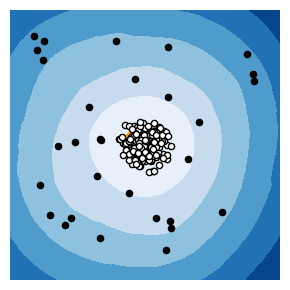}\\
        \hline
        \textbf{S2} & \includegraphics[width=0.2\textwidth]{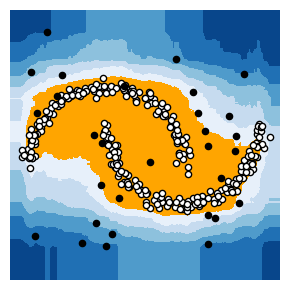} & \includegraphics[width=0.2\textwidth]{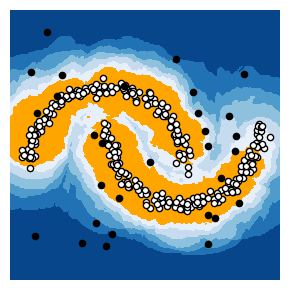}
        &\includegraphics[width=0.2\textwidth]{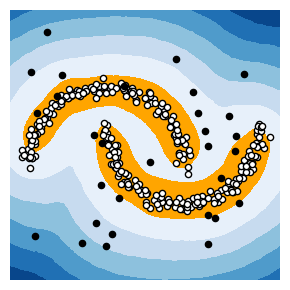}\\
        \hline
        \textbf{S3} & \includegraphics[width=0.2\textwidth]{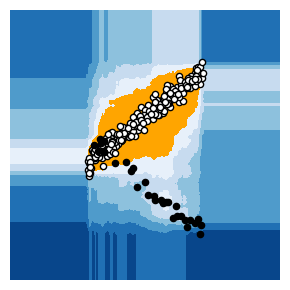} & \includegraphics[width=0.2\textwidth]{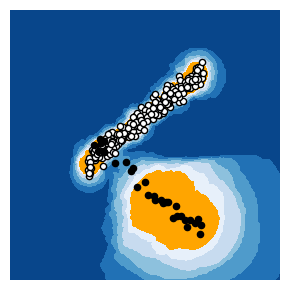}
        &\includegraphics[width=0.2\textwidth]{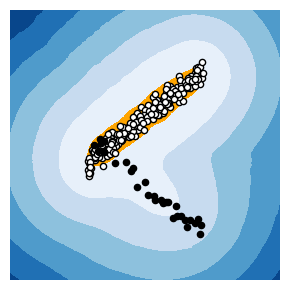}\\
    \hline
    \end{tabular}}
\end{table}

\subsection{Sensitivity and Scalability Analysis}

\begin{figure}[!ht]
     \centering
     \begin{subfigure}[b]{0.23\textwidth}
         \centering
         \includegraphics[width=1\textwidth]{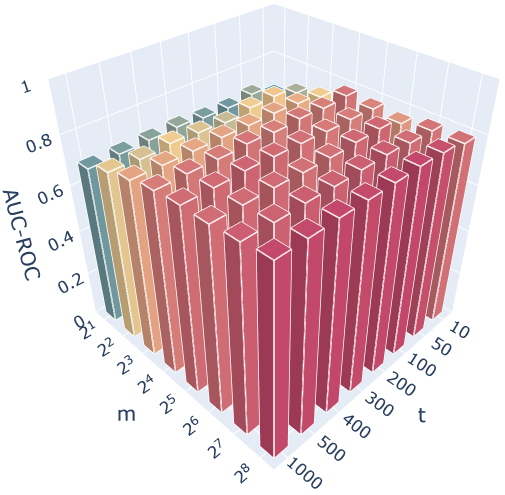}
         \caption{AUC-ROC}
         \label{fig:sensitivity_m}
     \end{subfigure}
     \begin{subfigure}[b]{0.23\textwidth}
         \centering
         \includegraphics[width=1\textwidth]{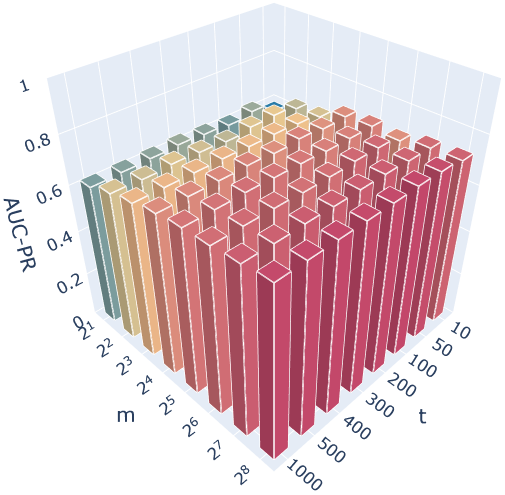}
         \caption{AUC-PR}
         \label{fig:sensitivity_t}
     \end{subfigure}
     \caption{The AUC-ROC and AUC-PR performance with different values of $m$ and $t$ across \texttt{magic.gamma} datasets.}
     \label{fig:sens}
\end{figure}

\textbf{Sensitivity Analysis:} We investigate SVEAD's sensitivity to two key hyperparameters: the number of sampled anchors $m$ per partition and the ensemble size $t$. Figure~\ref{fig:sens} shows AUC-ROC and AUC-PR performance across \texttt{magic.gamma} dataset. SVEAD shows sensitivity to the number of anchor points ($m$), as this directly determines the granularity of local region partitioning and affects the density adaptation mechanism through Voronoi cell formation. Larger $m$ values create finer partitions that capture more localized density variations, while smaller values may overlook subtle anomalies in complex data structures. In contrast, ensemble size $t$ demonstrates robustness due to variance reduction from ensemble averaging. The stochastic sampling of anchor points introduces variability across individual partitions, but this variance is effectively stabilized through aggregation, making the algorithm relatively insensitive to $t$. 

\begin{figure}[!ht]
     \centering
     \begin{subfigure}[b]{0.235\textwidth}
         \centering
         \includegraphics[width=1\textwidth]{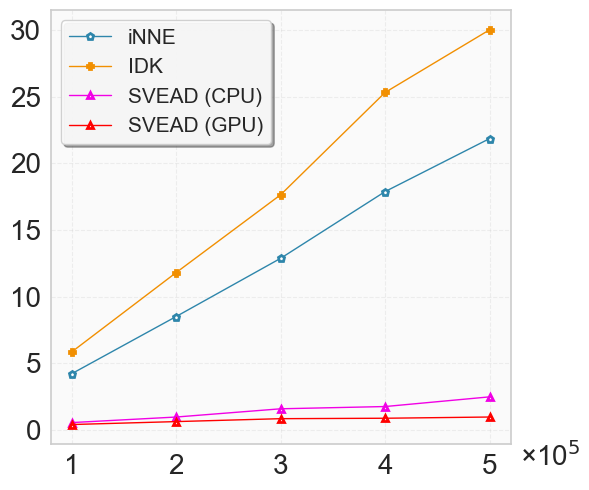}
         \caption{Samples}
         \label{fig:n_time}
     \end{subfigure}
     \begin{subfigure}[b]{0.235\textwidth}
         \centering
         \includegraphics[width=1\textwidth]{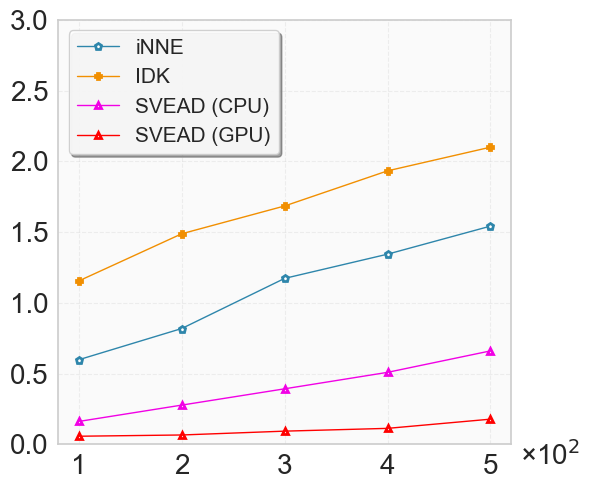}
         \caption{Dimensionalities}
         \label{fig:d_time}
     \end{subfigure}
     \caption{The runtime (seconds) comparison between SVEAD, iNNE and IDK. (a) fix dimensionality is 10, (b) fix samples are 1,0000. The test device is on CPU @AMD Ryzen 9 9950X and GPU @Nvidia RTX 4090.}
     \label{fig:time}
\end{figure}

\textbf{Scalability Analysis:} Figure~\ref{fig:time} compares the runtime of SVEAD with the two most competitive algorithms iNNE and IDK. Although all methods have linear time complexity, SVEAD demonstrates several-fold faster execution, requiring less than 3 seconds to process 500,000 instances on CPU. With GPU acceleration through parallel distance computation, SVEAD processes the same dataset in less than 1 second, providing substantial speedup for large-scale applications.

\vspace{-10pt}
\section{Conclusion}

We propose SVEAD that leverages stochastic Voronoi diagram and ensemble averaging for anomaly detection. The core contribution lies in recognizing that local anomalies, though indistinguishable under global analysis, become conspicuous when the data space is decomposed into restricted spatial regions. By randomly sampling anchors to induce Voronoi partitions and scoring points through normalized cell-relative distances weighted by local scale, SVEAD achieves density adaptation without explicit parameter tuning. Theoretical analysis establishes that random anchor sampling provides density-adaptive partitioning, ensemble averaging reduces variance and ensures robustness, and the dual-factor scoring mechanism creates clear separation between anomalous and normal instances. Extensive experiments on 45 benchmark datasets demonstrate that SVEAD achieves state-of-the-art performance with linear time complexity, outperforming 12 baselines. SVEAD's effectiveness across global, local, and dependency anomaly types, combined with its computational efficiency, makes it a practical solution for large-scale anomaly detection applications.

\section*{Impact Statement}




This paper aims to advance the field of anomaly detection in Machine Learning. SVEAD's computational efficiency and geometric interpretability could benefit applications in fraud detection, network security, industrial monitoring, and healthcare diagnostics, particularly in resource-constrained environments where real-time processing is critical. Like all anomaly detection systems, careful validation and appropriate human oversight remain essential when deployed in decision-making contexts, particularly in high-stakes domains where false positives or false negatives could have significant consequences. We encourage further research into diagnostic tools that help practitioners understand when geometric partitioning methods are most suitable for their specific applications, and into developing deployment guidelines that balance automation benefits with appropriate safeguards.

\bibliography{example_paper}
\bibliographystyle{icml2026}

\newpage
\appendix
\onecolumn



\section{Datasets}

\textbf{Synthetic datasets:} To assess SVEAD's ability to detect diverse anomaly patterns, we construct three synthetic datasets (300 samples each, 10\% anomaly rate). For \textbf{global anomalies} (S1), normal instances are sampled from a Gaussian distribution centered at the origin, while anomalies are uniformly scattered in peripheral sparse regions. For \textbf{local anomalies} (S2), normal instances form two dense moon-shaped clusters, and anomalies are randomly placed in the space between clusters where local density is low but not globally distant. For \textbf{dependency anomalies} (S3), normal instances follow a positive linear relationship, while anomalies exhibit a negative correlation, violating the feature dependency.

\textbf{Real-world datasets:} Table~\ref{tab:datasets} provides the statistic information of real-world datasets used in experiments from ADBench~\cite{han2022adbench}.

\begin{table}[!htbp]
\centering
\caption{Benchmark Datasets for Anomaly Detection.}
\label{tab:datasets}
\tiny
\resizebox{0.7\textwidth}{0.35\textheight}{
\begin{tabular}{lrrrrr}
\toprule
\textbf{Data} & \textbf{\# Samples} & \textbf{\# Features} & \textbf{\# Anomaly} & \textbf{\% Anomaly} & \textbf{Category} \\
\midrule
ALOI & 49534 & 27 & 1508 & 3.04 & Image \\
annthyroid & 7200 & 6 & 534 & 7.42 & Healthcare \\
backdoor & 95329 & 196 & 2329 & 2.44 & Network \\
breastw & 683 & 9 & 239 & 34.99 & Healthcare \\
campaign & 41188 & 62 & 4640 & 11.27 & Finance \\
cardio & 1831 & 21 & 176 & 9.61 & Healthcare \\
Cardio2 & 2114 & 21 & 466 & 22.04 & Healthcare \\
celeba & 202599 & 39 & 4547 & 2.24 & Image \\
census & 299285 & 500 & 18568 & 6.20 & Sociology \\
donors & 619326 & 10 & 36710 & 5.93 & Sociology \\
fault & 1941 & 27 & 673 & 34.67 & Physical \\
fraud & 284807 & 29 & 492 & 0.17 & Finance \\
glass & 214 & 7 & 9 & 4.21 & Forensic \\
Hepatitis & 80 & 19 & 13 & 16.25 & Healthcare \\
http & 567498 & 3 & 2211 & 0.39 & Web \\
InternetAds & 1966 & 1555 & 368 & 18.72 & Image \\
Ionosphere & 351 & 32 & 126 & 35.90 & Oryctognosy \\
landsat & 6435 & 36 & 1333 & 20.71 & Astronautics \\
letter & 1600 & 32 & 100 & 6.25 & Image \\
Lymphography & 148 & 18 & 6 & 4.05 & Healthcare \\
magic.gamma & 19020 & 10 & 6688 & 35.16 & Physical \\
mammography & 11183 & 6 & 260 & 2.32 & Healthcare \\
mnist & 7603 & 100 & 700 & 9.21 & Image \\
musk & 3062 & 166 & 97 & 3.17 & Chemistry \\
optdigits & 5216 & 64 & 150 & 2.88 & Image \\
PageBlocks & 5393 & 10 & 510 & 9.46 & Document \\
pendigits & 6870 & 16 & 156 & 2.27 & Image \\
Pima & 768 & 8 & 268 & 34.90 & Healthcare \\
satellite & 6435 & 36 & 2036 & 31.64 & Astronautics \\
satimage-2 & 5803 & 36 & 71 & 1.22 & Astronautics \\
shuttle & 49097 & 9 & 3511 & 7.15 & Astronautics \\
skin & 245057 & 3 & 50859 & 20.75 & Image \\
smtp & 95156 & 3 & 30 & 0.03 & Web \\
SpamBase & 4207 & 57 & 1679 & 39.91 & Document \\
speech & 3686 & 400 & 61 & 1.65 & Linguistics \\
Stamps & 340 & 9 & 31 & 9.12 & Document \\
thyroid & 3772 & 6 & 93 & 2.47 & Healthcare \\
vertebral & 240 & 6 & 30 & 12.50 & Biology \\
vowels & 1456 & 12 & 50 & 3.43 & Linguistics \\
Waveform & 3443 & 21 & 100 & 2.90 & Physics \\
WBC & 223 & 9 & 10 & 4.48 & Healthcare \\
WDBC & 367 & 30 & 10 & 2.72 & Healthcare \\
Wilt & 4819 & 5 & 257 & 5.33 & Botany \\
wine & 129 & 13 & 10 & 7.75 & Chemistry \\
WPBC & 198 & 33 & 47 & 23.74 & Healthcare \\
\bottomrule
\end{tabular}
}
\end{table}

\section{Empricial Evaluation.}

\begin{figure}[!htbp]
     \centering
     \begin{subfigure}[b]{0.7\textwidth}
         \centering
         \includegraphics[width=1\textwidth]{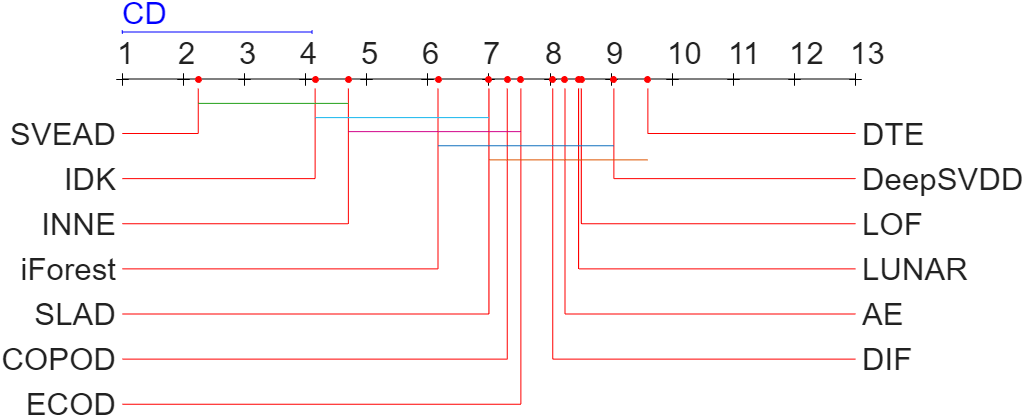}
         \caption{AUC-ROC}
     \end{subfigure}
     
     \begin{subfigure}[b]{0.7\textwidth}
         \centering
         \includegraphics[width=1\textwidth]{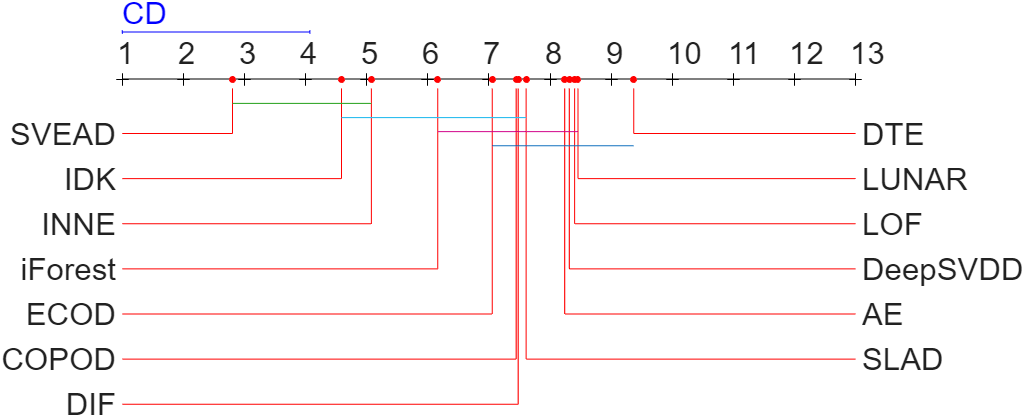}
         \caption{AUC-PR}
     \end{subfigure}
     \caption{Friedman-Nemenyi test for anomaly detectors at significance level 0.01. If two algorithms are connected by a CD (critical difference) line, there is no significant difference between them.}
     \label{fig:cd_diagram}
\end{figure}

\textbf{Significant analysis:} SVEAD achieves the highest average performance on both AUC-ROC and AUC-PR. Beyond average performance, we conduct rigorous statistical testing to assess whether SVEAD's improvements are significant. Figure~\ref{fig:cd_diagram} shows Friedman-Nemenyi test results, the critical difference (CD) diagrams reveal that SVEAD achieves the best rank among all methods on both AUC-ROC and AUC-PR metrics. Notably, SVEAD's rank is significantly better than traditional distance-based methods and most deep learning methods, demonstrating that the geometric partitioning approach combined with dual-factor scoring provides substantial advantages over existing paradigms. SVEAD is the only method that significant better than iForest and  other lower rank methods on both AUC-ROC and AUC-PR.

\begin{figure}[!htbp]
    \centering
    \includegraphics[width=0.5\linewidth]{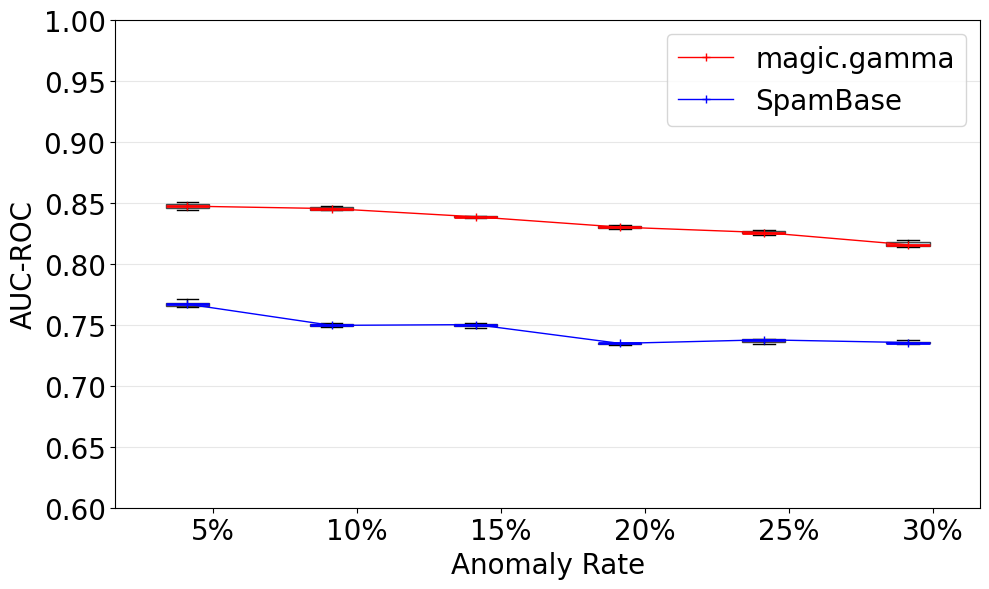}
    \caption{AUC-ROC of 10 runs on magic.gamma and SpamBase datasets with varying anomaly rates from 5\% to 30\%.}
    \label{fig:contamination}
\end{figure}

\textbf{Robustness to Contamination:} To evaluate SVEAD's robustness in the unsupervised setting, we test performance under varying contamination rates by randomly sampling different proportions of anomalies from the original datasets. Figure~\ref{fig:contamination} shows results on two datasets with anomaly rates ranging from 5\% to 30\%. SVEAD demonstrates stable performance across all contamination levels, the performance degradation remains minimal even as the anomaly rate increases to 30\%,. This robustness stems from SVEAD's scoring mechanism based on local relative positions and local density, which does not assume clean training data or require learning normal patterns. The method naturally adapts to contamination by identifying points that are consistently isolated across multiple random partitions.

\section{Ablation Study}

\begin{figure}[!htbp]
     \centering
     \begin{subfigure}[b]{0.3\textwidth}
         \centering
         \includegraphics[width=1\textwidth]{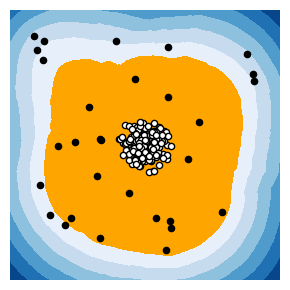}
         \caption{Global}
     \end{subfigure}
     \hfill
     \begin{subfigure}[b]{0.3\textwidth}
         \centering
         \includegraphics[width=1\textwidth]{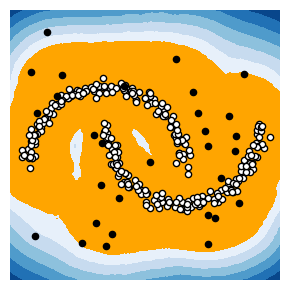}
         \caption{Local}
     \end{subfigure}
     \hfill
     \begin{subfigure}[b]{0.3\textwidth}
         \centering
         \includegraphics[width=1\textwidth]{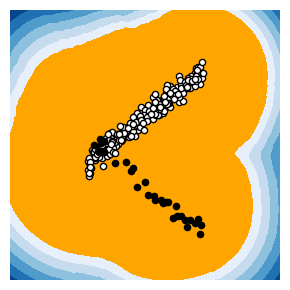}
         \caption{Dependency}
     \end{subfigure}
     \caption{Anomaly score heatmaps using only relative position term ($\delta/\delta_{\max}$) without local density weighting. }
     \label{fig:ablation}
\end{figure}

\textbf{Ablation Study:} To validate the necessity of the dual-factor scoring mechanism, we conduct ablation experiments using only the relative position term $\delta/\delta_{\max}$ without local density weighting $\delta_{\text{mean}}$. Figure~\ref{fig:ablation} shows results on three synthetic datasets. Without density weighting, the method loses its ability to distinguish anomalies from normal points. This occurs because the relative position term alone treats each cell independently, normalizing distances to [0,1] within every cell regardless of whether the cell is in a dense or sparse region. Consequently, boundary points in dense normal regions receive the same normalized scores as genuine anomalies in sparse regions, eliminating the critical density-based discrimination. The density weighting factor $\delta_{\text{mean}}$ addresses this by encoding local scale information: small $\delta_{\text{mean}}$ in dense regions suppresses scores, while large $\delta_{\text{mean}}$ in sparse regions amplifies scores, creating the necessary separation for effective anomaly detection. Conversely, using only $\delta_{\text{mean}}$ without relative position would fail to distinguish points within the same cell, highlighting the necessity of both factors.

\section{Full results}

Tables~\ref{tab:auc}, \ref{tab:auc-std}, \ref{tab:prc}, \ref{tab:prc-std} show mean and standard deviation of AUC-ROC and AUC-PR over 5 runs. Note that deterministic methods (e.g., ECOD, COPOD) have zero variance as they produce consistent outputs.

\begin{table}[!htbp]
\centering
\caption{Results in term of AUC-ROC $\uparrow$.}
\label{tab:auc}
\resizebox{\textwidth}{!}{%
\begin{tabular}{llllllllllllll}
\hline
\textbf{Data}                      & \textbf{LOF}    & \textbf{iForest} & \textbf{iNNE}   & \textbf{IDK}    & \textbf{ECOD}   & \textbf{COPOD}  & \textbf{D.SVDD} & \textbf{AE} & \textbf{LUNAR}  & \textbf{DIF} & \textbf{SLAD}   & \textbf{DTE}    & \textbf{VEAD}   \\ \hline
ALOI                               & \textbf{0.7668} & 0.5392           & 0.6506          & 0.6009          & 0.5305          & 0.5152          & 0.5395            & 0.5589               & {\ul 0.7058}    & 0.5490       & 0.5258          & 0.5256          & 0.6046          \\
annthyroid                         & 0.7392          & 0.8611           & 0.7542          & 0.7119          & 0.7887          & 0.7760          & 0.6522            & 0.7341               & 0.7303          & 0.6758       & {\ul 0.8696}    & 0.5593          & \textbf{0.9427} \\
backdoor                           & 0.7469          & 0.7355           & 0.8773          & 0.8626          & 0.8462          & 0.7893          & 0.8934            & 0.8900               & 0.5234          & {\ul 0.9215} & 0.9098          & 0.8709          & \textbf{0.9216} \\
breastw                            & 0.4687          & 0.9946           & 0.9843          & {\ul 0.9949}    & 0.9914          & 0.9944          & 0.9152            & 0.9635               & 0.9681          & 0.7989       & 0.6033          & 0.8699          & \textbf{0.9956} \\
campaign                           & 0.5982          & 0.7085           & 0.6818          & 0.6955          & 0.7697          & {\ul 0.7830}    & 0.6886            & 0.7548               & 0.6498          & 0.6779       & 0.7148          & 0.7099          & \textbf{0.8319} \\
cardio                             & 0.6179          & 0.9260           & {\ul 0.9447}    & \textbf{0.9531} & 0.9350          & 0.9219          & 0.9001            & 0.7507               & 0.6089          & 0.9281       & 0.4975          & 0.7651          & 0.9339          \\
Cardio2                            & 0.6271          & 0.6914           & {\ul 0.7907}    & 0.7701          & 0.7853          & 0.6629          & 0.6893            & 0.5644               & 0.5252          & 0.6211       & 0.6765          & 0.4703          & \textbf{0.8470} \\
celeba                             & 0.4786          & 0.7188           & 0.7429          & {\ul 0.7762}    & 0.7572          & 0.7508          & 0.7397            & 0.7253               & 0.5624          & 0.6493       & 0.7436          & \textbf{0.8034} & 0.7591          \\
census                             & 0.5761          & 0.6138           & 0.5755          & 0.6202          & 0.6595          & {\ul 0.6740}    & 0.6442            & 0.6722               & 0.6445          & 0.5784       & 0.5825          & 0.5448          & \textbf{0.6869} \\
donors                             & 0.6153          & 0.8476           & 0.8193          & 0.8777          & \textbf{0.8885} & 0.8151          & 0.8016            & 0.8325               & 0.6066          & 0.6873       & 0.5485          & 0.8072          & {\ul 0.8790}     \\
fault                              & 0.5973          & 0.5608           & 0.6146          & 0.6912          & 0.4687          & 0.4553          & 0.5170            & 0.6489               & {\ul 0.7202}    & 0.6889       & 0.7131          & 0.5873          & \textbf{0.7543} \\
fraud                              & 0.4899          & 0.9518           & \textbf{0.9575} & {\ul 0.9525}    & 0.9496          & 0.9475          & 0.9337            & 0.9431               & 0.9276          & 0.9503       & 0.9401          & 0.9405          & 0.9473          \\
glass                              & 0.8466          & 0.7883           & 0.8246          & 0.8481          & 0.7046          & 0.7550          & 0.6249            & 0.8108               & 0.8502          & 0.8507       & \textbf{0.8900}          & 0.6133          & {\ul 0.8697}    \\
Hepatitis                          & 0.5890          & 0.7201           & 0.5747          & 0.5711          & 0.7394          & \textbf{0.8037}          & 0.6429            & 0.7463               & 0.5931          & 0.7137       & 0.5844          & 0.3823          & {\ul 0.7897}    \\
http                               & 0.4392          & 0.9998           & 0.9968          & 0.9987          & 0.9786          & 0.9915          & 0.9972            & 0.9747               & 0.2614          & 0.9933       & 0.9992          & 0.9963          & \textbf{1.0000} \\
InternetADs                        & 0.6470          & 0.6828           & {\ul 0.6954}    & 0.6891          & 0.6770          & 0.6764          & 0.6045            & 0.5593               & \textbf{0.6960} & 0.5522       & 0.6158          & 0.5901          & 0.6935          \\
Ionosphere                         & 0.8944          & 0.8503           & 0.8910          & 0.8861          & 0.7284          & 0.7895          & 0.7246            & 0.8909               & 0.9182          & 0.9015       & \textbf{0.9414} & 0.8869          & {\ul 0.9315}    \\
landsat                            & 0.5466          & 0.4666           & 0.6148          & \textbf{0.6992} & 0.3678          & 0.4215          & 0.3918            & 0.5107               & 0.5714          & 0.5583       & {\ul 0.6667}    & 0.5242          & 0.6021          \\
letter                             & 0.9131          & 0.6179           & {\ul 0.9213}    & \textbf{0.9216} & 0.5326          & 0.5032          & 0.5167            & 0.8365               & 0.8880          & 0.6474       & 0.7394          & 0.7774          & 0.8949          \\
Lymphography                       & 0.9906          & \textbf{0.9991}  & 0.9948          & 0.9901          & 0.9953          & 0.9965          & 0.9862            & 0.9812               & 0.9596          & 0.9202       & 0.9636          & 0.9688          & {\ul 0.9986}    \\
{magic.gamma} & 0.7110          & 0.7240           & 0.7491          & 0.7584          & 0.6382          & 0.6812          & 0.6282            & 0.7409               & {\ul 0.8192}    & 0.7587       & 0.6243          & 0.7590          & \textbf{0.8241} \\
{mammography} & 0.7351          & 0.8659           & 0.8659          & 0.8719          & \textbf{0.9062} & {\ul 0.9053}    & 0.8819            & 0.8351               & 0.8390          & 0.7450       & 0.6929          & 0.8265          & 0.8749          \\
{mnist}       & 0.7133          & 0.8043           & 0.8384          & 0.8370          & 0.7463          & 0.7739          & 0.7951            & 0.8369               & 0.7416          & {\ul 0.8699} & \textbf{0.9302} & 0.5424          & 0.8664          \\
{musk}        & 0.4912          & {\ul 0.9997}     & \textbf{1.0000} & \textbf{1.0000} & 0.9559          & 0.9463          & 0.9466            & 0.8562               & 0.3118          & 0.9819       & 0.8706          & 0.5599          & \textbf{1.0000} \\
{optdigits}   & 0.6660          & 0.7352           & {\ul 0.8338}    & \textbf{0.9019} & 0.6045          & 0.6824          & 0.4551            & 0.4781               & 0.4243          & 0.5670       & 0.5751          & 0.5000          & 0.7035          \\
{PageBlocks}  & 0.8328          & 0.8974           & 0.8958          & 0.8561          & {\ul 0.9139}    & 0.8754          & 0.9067            & 0.8789               & 0.7631          & 0.8811       & 0.9109          & 0.4860          & \textbf{0.9533} \\
{pendigits}   & 0.5230          & 0.9542           & 0.9291          & 0.9581          & 0.9274          & 0.9048          & 0.7582            & 0.8416               & 0.6938          & 0.9440       & \textbf{0.9624} & 0.8864          & {\ul 0.9603}    \\
{Pima}        & 0.5995          & {\ul 0.6957}     & 0.6493          & 0.6389          & 0.5944          & 0.6540          & 0.6773            & 0.6278               & 0.6886          & 0.6122       & 0.4812          & 0.4455          & \textbf{0.7311} \\
{satellite}   & 0.5637          & 0.6996           & 0.7465          & \textbf{0.8046} & 0.5830          & 0.6335          & 0.6111            & 0.6658               & 0.6380          & 0.7395       & 0.6270          & 0.5832          & {\ul 0.7905}    \\
{satimage-2}  & 0.5765          & 0.9948           & {\ul 0.9978}    & 0.9975          & 0.9649          & 0.9745          & 0.964             & 0.9033               & 0.8993          & 0.9969       & 0.9615          & 0.496           & \textbf{0.9988} \\
{shuttle}     & 0.5819          & \textbf{0.9972}  & 0.9942          & 0.9934          & 0.9929          & 0.9945          & 0.9883            & 0.9331               & 0.6725          & 0.9672       & 0.9033          & 0.5             & {\ul 0.9951}    \\
{skin}        & 0.5537          & 0.6713           & 0.7535          & {\ul 0.8075}    & 0.4888          & 0.4711          & 0.5333            & 0.6827               & 0.7156          & 0.6688       & \textbf{0.8390} & 0.7662          & 0.7773          \\
{smtp}        & 0.9270          & 0.9173           & {\ul 0.9460}    & \textbf{0.9565} & 0.8801          & 0.9120          & 0.8444            & 0.8987               & 0.9215          & 0.8521       & 0.8122          & 0.9157          & 0.9296          \\
{SpamBase}    & 0.5079          & 0.6981           & 0.7138          & {\ul 0.7245}    & 0.6556          & 0.6879          & 0.5589            & 0.5533               & 0.5205          & 0.3901       & 0.3985          & 0.5629          & \textbf{0.7376} \\
{speech}      & 0.6494          & 0.4828           & 0.6730          & \textbf{0.7389} & 0.4697          & 0.4911          & 0.5035            & 0.4751               & 0.4970          & 0.4896       & 0.5463          & 0.5270          & {\ul 0.6919}    \\
{Stamps}      & 0.7076          & {\ul 0.9253}     & 0.8900          & 0.9114          & 0.8761          & \textbf{0.9302} & 0.8422            & 0.8277               & 0.8456          & 0.8589       & 0.8354          & 0.8320          & 0.9064          \\
{thyroid}     & 0.9287          & {\ul 0.9872}     & 0.9566          & 0.9577          & 0.9771          & 0.9393          & 0.9295            & 0.9238               & 0.9361          & 0.9595       & 0.9654          & 0.9029          & \textbf{0.9915} \\
{vertebral}   & 0.5298          & 0.3770           & 0.5640          & 0.5579          & 0.4200          & 0.3349          & 0.3413            & 0.4660               & 0.4001          & 0.4972       & 0.4176          & {\ul 0.6728}    & \textbf{0.7345} \\
{vowels}      & 0.9434          & 0.7736           & 0.9392          & {\ul 0.9542}    & 0.5929          & 0.4958          & 0.6164            & 0.9385               & 0.9220          & 0.8116       & 0.9456          & 0.9334          & \textbf{0.9802} \\
{Waveform}    & 0.7632          & 0.7147           & 0.8410          & {\ul 0.8679}    & 0.6035          & 0.7339          & 0.5648            & 0.6085               & 0.7300          & 0.7255       & \textbf{0.8842} & 0.6316          & 0.8345          \\
{WBC}         & 0.9685          & {\ul 0.9963}     & 0.9955          & 0.9947          & 0.9948          & 0.9948          & 0.9924            & 0.9587               & 0.9630          & 0.8130       & 0.8606          & 0.5716          & \textbf{0.9975} \\
{WDBC}        & 0.9989          & 0.9888           & 0.9987          & {\ul 0.9990}    & 0.9706          & 0.9941          & 0.9842            & 0.9300               & 0.9593          & 0.7087       & 0.9986          & 0.5831          & \textbf{0.9992} \\
{Wilt}        & \textbf{0.7750} & 0.4625           & {\ul 0.7683}    & 0.7572          & 0.3940          & 0.3447          & 0.3449            & 0.5634               & 0.5159          & 0.3788       & 0.4915          & 0.4689          & 0.7177          \\
{wine}        & \textbf{0.9992} & 0.8304           & 0.9988          & {\ul 0.9991}    & 0.7328          & 0.8672          & 0.7368            & 0.7613               & 0.3753          & 0.5139       & 0.9210          & 0.3820          & 0.9987          \\
{WPBC}        & 0.5185          & 0.5075           & 0.5124          & 0.5167          & 0.4813          & 0.5233          & 0.4658            & 0.4753               & 0.4947          & {\ul 0.4690} & 0.5359          & 0.4011          & \textbf{0.5414} \\ \hline
{\textbf{Average}}     & 0.6879          & 0.7772           & 0.8213          & {\ul 0.8327}          & 0.7435          & 0.7504          & 0.7172            & 0.7558               & 0.6933          & 0.7348       & 0.7493          & 0.6651          & \textbf{0.8538} \\ \hline
\end{tabular}%
}
\end{table}

\begin{table}[!htbp]
\centering
\caption{Results in term of AUC-ROC standard deviation.}
\label{tab:auc-std}
\resizebox{\textwidth}{!}{%
\begin{tabular}{llllllllllllll}
\hline
\textbf{Data}                      & \textbf{LOF}    & \textbf{iForest} & \textbf{iNNE}   & \textbf{IDK}    & \textbf{ECOD}   & \textbf{COPOD}  & \textbf{D.SVDD} & \textbf{AE} & \textbf{LUNAR}  & \textbf{DIF} & \textbf{SLAD}   & \textbf{DTE}    & \textbf{VEAD}   \\ \hline
ALOI                               & 0.0000       & 0.0010           & 0.0036        & 0.0045       & 0.0000        & 0.0000         & 0.0081            & 0.0000               & 0.0061         & 0.0007       & 0.0000        & 0.0003       & 0.0007         \\
annthyroid                         & 0.0000       & 0.0081           & 0.0030        & 0.0060       & 0.0000        & 0.0000         & 0.0114            & 0.0000               & 0.0156         & 0.0044       & 0.0000        & 0.0155       & 0.0024         \\
backdoor                           & 0.0000       & 0.0267           & 0.0072        & 0.0233       & 0.0000        & 0.0000         & 0.0121            & 0.0000               & 0.0426         & 0.0042       & 0.0000        & 0.0191       & 0.0030         \\
breastw                            & 0.0000       & 0.0008           & 0.0031        & 0.0002       & 0.0000        & 0.0000         & 0.0662            & 0.0000               & 0.0084         & 0.0242       & 0.0000        & 0.0198       & 0.0001         \\
campaign                           & 0.0000       & 0.0120           & 0.0040        & 0.0025       & 0.0000        & 0.0000         & 0.0343            & 0.0000               & 0.0148         & 0.0113       & 0.0000        & 0.0118       & 0.0049         \\
cardio                             & 0.0000       & 0.0125           & 0.0035        & 0.0033       & 0.0000        & 0.0000         & 0.0457            & 0.0000               & 0.0259         & 0.0081       & 0.0000        & 0.0437       & 0.0063         \\
Cardio2                            & 0.0000       & 0.0194           & 0.0156        & 0.0140       & 0.0000        & 0.0000         & 0.0544            & 0.0000               & 0.0055         & 0.0127       & 0.0000        & 0.0904       & 0.0148         \\
celeba                             & 0.0000       & 0.0148           & 0.0155        & 0.0173       & 0.0000        & 0.0000         & 0.0343            & 0.0000               & 0.0037         & 0.0070       & 0.0000        & 0.0385       & 0.0244         \\
census                             & 0.0000       & 0.0285           & 0.0439        & 0.0107       & 0.0000        & 0.0000         & 0.0159            & 0.0000               & 0.0029         & 0.0107       & 0.0000        & 0.0303       & 0.0050         \\
donors                             & 0.0000       & 0.0235           & 0.0302        & 0.0394       & 0.0000        & 0.0000         & 0.0855            & 0.0000               & 0.0004         & 0.0148       & 0.0000        & 0.0384       & 0.0018         \\
fault                              & 0.0000       & 0.0217           & 0.0124        & 0.0099       & 0.0000        & 0.0000         & 0.0473            & 0.0000               & 0.0066         & 0.0087       & 0.0000        & 0.0326       & 0.0053         \\
fraud                              & 0.0000       & 0.0020           & 0.0007        & 0.0019       & 0.0000        & 0.0000         & 0.0046            & 0.0000               & 0.0062         & 0.9503       & 0.0000        & 0.0065       & 0.0009         \\
glass                              & 0.0000       & 0.0154           & 0.0232        & 0.0127       & 0.0000        & 0.0000         & 0.1178            & 0.0000               & 0.0264         & 0.0151       & 0.0000        & 0.0484       & 0.0091         \\
Hepatitis                          & 0.0000       & 0.0276           & 0.0384        & 0.0202       & 0.0000        & 0.0000         & 0.1044            & 0.0000               & 0.0690         & 0.0351       & 0.0000        & 0.0583       & 0.0097         \\
http                               & 0.0000       & 0.0003           & 0.0028        & 0.0005       & 0.0000        & 0.0000         & 0.0012            & 0.0000               & 0.0009         & 0.0000       & 0.0000        & 0.0019       & 0.0000         \\
InternetADs                        & 0.0000       & 0.0230           & 0.0028        & 0.0035       & 0.0000        & 0.0000         & 0.0305            & 0.0000               & 0.0040         & 0.0245       & 0.0000        & 0.0642       & 0.0037         \\
Ionosphere                         & 0.0000       & 0.0032           & 0.0056        & 0.0065       & 0.0000        & 0.0000         & 0.0383            & 0.0000               & 0.0123         & 0.0037       & 0.0000        & 0.0176       & 0.0020         \\
landsat                            & 0.0000       & 0.0053           & 0.0082        & 0.0245       & 0.0000        & 0.0000         & 0.0540            & 0.0000               & 0.0019         & 0.0037       & 0.0000        & 0.0326       & 0.0055         \\
letter                             & 0.0000       & 0.0165           & 0.0047        & 0.0069       & 0.0000        & 0.0000         & 0.0336            & 0.0000               & 0.0128         & 0.0306       & 0.0000        & 0.0302       & 0.0095         \\
Lymphography                       & 0.0000       & 0.0014           & 0.0034        & 0.0009       & 0.0000        & 0.0000         & 0.0143            & 0.0000               & 0.0669         & 0.0197       & 0.0000        & 0.0079       & 0.0005         \\
magic.gamma & 0.0000       & 0.0088           & 0.0138        & 0.0129       & 0.0000        & 0.0000         & 0.0146            & 0.0000               & 0.0040         & 0.0055       & 0.0000        & 0.0107       & 0.0009         \\
mammography & 0.0000       & 0.0092           & 0.0089        & 0.0031       & 0.0000        & 0.0000         & 0.0138            & 0.0000               & 0.0047         & 0.0183       & 0.0000        & 0.0349       & 0.0066         \\
mnist       & 0.0000       & 0.0185           & 0.0177        & 0.0070       & 0.0000        & 0.0000         & 0.0874            & 0.0000               & 0.0439         & 0.0055       & 0.0000        & 0.0849       & 0.0036         \\
musk        & 0.0000       & 0.0004           & 0.0000        & 0.0000       & 0.0000        & 0.0000         & 0.0757            & 0.0000               & 0.1507         & 0.0068       & 0.0000        & 0.2352       & 0.0000         \\
optdigits   & 0.0000       & 0.0395           & 0.0279        & 0.0218       & 0.0000        & 0.0000         & 0.1243            & 0.0000               & 0.0376         & 0.0543       & 0.0000        & 0.0000       & 0.0546         \\
PageBlocks  & 0.0000       & 0.0041           & 0.0088        & 0.0346       & 0.0000        & 0.0000         & 0.0150            & 0.0000               & 0.0027         & 0.0087       & 0.0000        & 0.1471       & 0.0037         \\
pendigits   & 0.0000       & 0.0046           & 0.0102        & 0.0083       & 0.0000        & 0.0000         & 0.1223            & 0.0000               & 0.0073         & 0.0064       & 0.0000        & 0.0091       & 0.0058         \\
Pima        & 0.0000       & 0.0182           & 0.0099        & 0.0232       & 0.0000        & 0.0000         & 0.0319            & 0.0000               & 0.0145         & 0.0105       & 0.0000        & 0.0722       & 0.0033         \\
satellite   & 0.0000       & 0.0049           & 0.0098        & 0.0169       & 0.0000        & 0.0000         & 0.0450            & 0.0000               & 0.0012         & 0.0068       & 0.0000        & 0.0704       & 0.0045         \\
satimage-2  & 0.0000       & 0.0009           & 0.0002        & 0.0004       & 0.0000        & 0.0000         & 0.0118            & 0.0000               & 0.0287         & 0.0004       & 0.0000        & 0.0706       & 0.0001         \\
shuttle     & 0.0000       & 0.0006           & 0.0001        & 0.0003       & 0.0000        & 0.0000         & 0.0016            & 0.0000               & 0.0105         & 0.0068       & 0.0000        & 0.0000       & 0.0009         \\
skin        & 0.0000       & 0.0100           & 0.0359        & 0.0225       & 0.0000        & 0.0000         & 0.0377            & 0.0000               & 0.0007         & 0.0072       & 0.0000        & 0.0149       & 0.0122         \\
smtp        & 0.0000       & 0.0053           & 0.0067        & 0.0031       & 0.0000        & 0.0000         & 0.0191            & 0.0000               & 0.0028         & 0.0026       & 0.0000        & 0.0053       & 0.0010         \\
SpamBase    & 0.0000       & 0.0293           & 0.0168        & 0.0062       & 0.0000        & 0.0000         & 0.0350            & 0.0000               & 0.0170         & 0.0175       & 0.0000        & 0.1642       & 0.0010         \\
speech      & 0.0000       & 0.0067           & 0.0083        & 0.0027       & 0.0000        & 0.0000         & 0.0381            & 0.0000               & 0.0254         & 0.0334       & 0.0000        & 0.0336       & 0.0150         \\
Stamps      & 0.0000       & 0.0029           & 0.0145        & 0.0043       & 0.0000        & 0.0000         & 0.0656            & 0.0000               & 0.0217         & 0.0166       & 0.0000        & 0.0240       & 0.0087         \\
thyroid     & 0.0000       & 0.0040           & 0.0054        & 0.0022       & 0.0000        & 0.0000         & 0.0119            & 0.0000               & 0.0107         & 0.0030       & 0.0000        & 0.0046       & 0.0002         \\
vertebral   & 0.0000       & 0.0385           & 0.0000        & 0.0144       & 0.0000        & 0.0000         & 0.0446            & 0.0000               & 0.0254         & 0.0230       & 0.0000        & 0.1015       & 0.0150         \\
vowels      & 0.0000       & 0.0210           & 0.0048        & 0.0051       & 0.0000        & 0.0000         & 0.1495            & 0.0000               & 0.0447         & 0.0173       & 0.0000        & 0.0133       & 0.0020         \\
Waveform    & 0.0000       & 0.0255           & 0.0158        & 0.0106       & 0.0000        & 0.0000         & 0.0810            & 0.0000               & 0.0109         & 0.0302       & 0.0000        & 0.0262       & 0.0082         \\
WBC         & 0.0000       & 0.0007           & 0.0021        & 0.0011       & 0.0000        & 0.0000         & 0.0028            & 0.0000               & 0.0333         & 0.0225       & 0.0000        & 0.0583       & 0.0004         \\
WDBC        & 0.0000       & 0.0014           & 0.0011        & 0.0004       & 0.0000        & 0.0000         & 0.0082            & 0.0000               & 0.0127         & 0.0381       & 0.0000        & 0.4754       & 0.0003         \\
Wilt        & 0.0000       & 0.0250           & 0.0057        & 0.0098       & 0.0000        & 0.0000         & 0.0361            & 0.0000               & 0.0072         & 0.0226       & 0.0000        & 0.1535       & 0.0030         \\
wine        & 0.0000       & 0.0287           & 0.0002        & 0.0003       & 0.0000        & 0.0000         & 0.2260            & 0.0000               & 0.0554         & 0.0709       & 0.0000        & 0.2412       & 0.0004         \\
WPBC        & 0.0000       & 0.0429           & 0.0091        & 0.0133       & 0.0000        & 0.0000         & 0.0161            & 0.0000               & 0.0197         & 0.0136       & 0.0000        & 0.0202       & 0.0073        
\\ \hline
\end{tabular}%
}
\end{table}

\begin{table}[]
\caption{Results in term of AUC-PR $\uparrow$.}
\label{tab:prc}
\resizebox{\textwidth}{!}{%
\begin{tabular}{llllllllllllll}
\hline
\textbf{Data}                      & \textbf{LOF}    & \textbf{iForest} & \textbf{iNNE}   & \textbf{IDK}    & \textbf{ECOD}   & \textbf{COPOD}  & \textbf{D.SVDD} & \textbf{AE} & \textbf{LUNAR}  & \textbf{DIF} & \textbf{SLAD}   & \textbf{DTE} & \textbf{VEAD} \\ \hline
ALOI                               & \textbf{0.1033} & 0.0337           & 0.0894          & 0.0623          & 0.0329          & 0.0313          & 0.0369            & 0.0404               & {\ul 0.0906}    & 0.0427       & 0.0444          & 0.0436       & 0.0683             \\
annthyroid                         & 0.2053          & 0.3963           & 0.2043          & 0.2001          & 0.2697          & 0.1733          & 0.1766            & 0.2178               & 0.1834          & 0.2196       & {\ul 0.4167}    & 0.1142       & \textbf{0.6050}    \\
backdoor                           & 0.2075          & 0.0459           & 0.1310          & 0.1059          & 0.0923          & 0.0685          & {\ul 0.5821}      & 0.4967               & 0.0481          & 0.4099       & 0.5390          & 0.3900       & \textbf{0.5925}    \\
breastw                            & 0.3064          & 0.9893           & 0.9618          & {\ul 0.9900}    & 0.9839          & 0.9886          & 0.9189            & 0.9010               & 0.9192          & 0.4900       & 0.5509          & 0.6715       & \textbf{0.9914}    \\
campaign                           & 0.1344          & 0.2955           & 0.2181          & 0.2160          & {\ul 0.3546}    & \textbf{0.3686} & 0.2470            & 0.2470               & 0.1932          & 0.2441       & 0.2888          & 0.2613       & 0.3267             \\
cardio                             & 0.1729          & 0.5861           & 0.5684          & \textbf{0.6381} & 0.5674          & 0.5776          & 0.5231            & 0.3038               & 0.2066          & 0.5909       & 0.1653          & 0.2948       & 0.5356             \\
Cardio2                   & 0.3196          & 0.4348           & 0.4645          & 0.4310          & {\ul 0.5054}    & 0.4001          & 0.4053            & 0.3133               & 0.2784          & 0.4171       & 0.3426          & 0.2378       & \textbf{0.5567}    \\
celeba                             & 0.0202          & 0.0694           & 0.0805          & 0.0867          & \textbf{0.0969} & {\ul 0.0944}    & 0.0817            & 0.0542               & 0.0262          & 0.0463       & 0.0461          & 0.0677       & 0.0835             \\
census                             & 0.0750          & 0.0742           & 0.0692          & 0.0736          & 0.0840          & 0.0877          & 0.0834            & \textbf{0.0909}      & 0.0818          & 0.0672       & 0.0682          & 0.0657       & {\ul 0.0882}       \\
cover                              & 0.0175          & 0.0459           & \textbf{0.3059} & {\ul 0.2988}    & 0.1072          & 0.0658          & 0.0606            & 0.0700               & 0.0376          & 0.2025       & 0.0135          & 0.0189       & 0.0569             \\
donors                             & 0.1039          & 0.1947           & 0.1571          & 0.2066          & \textbf{0.2641} & {\ul 0.2082}    & 0.1603            & 0.1686               & 0.1017          & 0.0825       & 0.0597          & 0.1414       & 0.2057             \\
fault                              & 0.4020          & 0.4038           & 0.4693          & 0.5062          & 0.3257          & 0.3127          & 0.3685            & 0.4824               & {\ul 0.5365}    & 0.5361       & 0.5291          & 0.4211       & \textbf{0.5664}    \\
fraud                              & 0.0016          & 0.2010           & 0.1541          & 0.1147          & 0.2167          & 0.2431          & 0.1158            & 0.1495               & 0.1075          & {\ul 0.3700} & 0.1274          & 0.2925       & \textbf{0.4176}    \\
glass                              & \textbf{0.2539} & 0.1150           & 0.1535          & 0.1433          & 0.1859          & 0.1085          & 0.1067            & 0.1655               & 0.1796          & {\ul 0.2347} & 0.1817          & 0.0793       & 0.2089             \\
Hepatitis                          & 0.2640          & 0.2736           & 0.2215          & 0.2310          & 0.2934          & {\ul 0.3941}    & 0.2785            & 0.3264               & 0.1963          & 0.2971       & 0.1891          & 0.1486       & \textbf{0.4190}    \\
http                               & 0.0070          & {\ul 0.9513}     & 0.6378          & 0.7092          & 0.1453          & 0.3018          & 0.5802            & 0.1246               & 0.0004          & 0.3505       & 0.8236          & 0.5517       & \textbf{0.9985}    \\
InternetADs                        & 0.4070          & 0.4655           & {\ul 0.5203}    & \textbf{0.5222} & 0.5089          & 0.5077          & 0.2697            & 0.2092               & 0.3879          & 0.2047       & 0.2688          & 0.2914       & 0.3861             \\
Ionosphere                         & 0.8645          & 0.8013           & 0.8798          & 0.8611          & 0.6461          & 0.6697          & 0.5959            & 0.8619               & \textbf{0.9186} & 0.8722       & 0.9011          & 0.8536       & {\ul 0.9173}       \\
landsat                            & 0.2516          & 0.1960           & 0.2714          & \textbf{0.3332} & 0.1635          & 0.1758          & 0.1922            & 0.2231               & 0.2493          & 0.2514       & {\ul 0.3021}    & 0.2153       & 0.2491             \\
letter                             & \textbf{0.6071} & 0.0850           & 0.5722          & {\ul 0.5754}    & 0.0757          & 0.0730          & 0.0690            & 0.2795               & 0.3810          & 0.1114       & 0.1628          & 0.3104       & 0.3618             \\
Lymphography                       & 0.8215          & \textbf{0.9813}  & 0.9001          & 0.7510          & 0.8972          & 0.9151          & 0.7839            & 0.7306               & 0.8485          & 0.4422       & 0.5099          & 0.4774       & {\ul 0.9704}       \\
{magic.gamma} & 0.5560          & 0.6399           & 0.6478          & 0.6318          & 0.5335          & 0.5881          & 0.5538            & 0.6399               & {\ul 0.7492}    & 0.6761       & 0.4740          & 0.6607       & \textbf{0.7544}    \\
{mammography} & 0.1160          & 0.2163           & 0.1795          & 0.1892          & \textbf{0.4360} & {\ul 0.4300}    & 0.2149            & 0.1171               & 0.1486          & 0.1213       & 0.0567          & 0.1829       & 0.1869             \\
{mnist}       & 0.2426          & 0.2705           & 0.3370          & 0.3270          & 0.1780          & 0.2140          & 0.3434            & 0.3610               & 0.2980          & {\ul 0.4038} & \textbf{0.5749} & 0.1188       & 0.3836             \\
{musk}        & 0.1283          & {\ul 0.9902}     & \textbf{1.0000} & \textbf{1.0000} & 0.4918          & 0.3433          & 0.6828            & 0.4097               & 0.0453          & 0.7202       & 0.2283          & 0.0521       & \textbf{1.0000}    \\
{optdigits}   & 0.0545          & 0.0544           & {\ul 0.0810}    & \textbf{0.1357} & 0.0337          & 0.0422          & 0.0256            & 0.0255               & 0.0264          & 0.0323       & 0.0322          & 0.0288       & 0.0473             \\
{PageBlocks}  & 0.4875          & 0.4655           & 0.5501          & {\ul 0.5667}    & 0.5199          & 0.3710          & 0.5664            & 0.3944               & 0.3655          & 0.5064       & 0.4937          & 0.1825       & \textbf{0.6769}    \\
{pendigits}   & 0.0486          & 0.2974           & 0.2262          & {\ul 0.3056}    & 0.2656          & 0.1824          & 0.1157            & 0.0797               & 0.0544          & 0.2441       & \textbf{0.3475} & 0.0987       & 0.2837             \\
{Pima}        & 0.4324          & \textbf{0.5422}  & 0.4970          & 0.4912          & 0.4642          & 0.5054          & 0.5126            & 0.4458               & 0.5050          & 0.4124       & 0.3633          & 0.3674       & {\ul 0.5334}       \\
{satellite}   & 0.3916          & 0.6733           & 0.6194          & {\ul 0.6851}    & 0.5261          & 0.5706          & 0.5633            & 0.5504               & 0.4785          & 0.6795       & 0.4181          & 0.3964       & \textbf{0.7126}    \\
{satimage-2}  & 0.0634          & 0.9348           & {\ul 0.9654}    & 0.8279          & 0.6597          & 0.7933          & 0.6428            & 0.2985               & 0.1578          & 0.7737       & 0.3576          & 0.0120       & \textbf{0.9723}    \\
{shuttle}     & 0.1275          & \textbf{0.9798}  & 0.8840          & 0.8511          & 0.9043          & {\ul 0.9608}    & 0.9059            & 0.6271               & 0.1794          & 0.5897       & 0.3544          & 0.0715       & 0.9005             \\
{skin}        & 0.2353          & 0.2546           & 0.3166          & {\ul 0.3747}    & 0.1821          & 0.1783          & 0.2045            & 0.2621               & 0.2911          & 0.2551       & \textbf{0.4854} & 0.3236       & 0.3367             \\
{smtp}        & 0.0272          & 0.0775           & 0.3888          & 0.2803          & {\ul 0.5073}    & 0.0041          & 0.2267            & 0.0035               & 0.0582          & 0.4560       & 0.0224          & 0.3828       & \textbf{0.5337}    \\
{SpamBase}    & 0.3952          & 0.5631           & 0.6195          & {\ul 0.6299}    & 0.5183          & 0.5438          & 0.4181            & 0.4150               & 0.4160          & 0.3331       & 0.3436          & 0.4964       & \textbf{0.6338}    \\
{speech}      & 0.0703          & 0.0188           & \textbf{0.1180} & {\ul 0.1155}    & 0.0196          & 0.0193          & 0.0189            & 0.0186               & 0.0197          & 0.0195       & 0.0194          & 0.0200       & 0.0820             \\
{Stamps}      & 0.2018          & {\ul 0.3753}     & 0.3268          & 0.3716          & 0.3142          & \textbf{0.3961} & 0.2767            & 0.3049               & 0.3152          & 0.2737       & 0.2954          & 0.2569       & 0.3594             \\
{thyroid}     & 0.2073          & {\ul 0.6536}     & 0.3110          & 0.3507          & 0.4678          & 0.1789          & 0.2859            & 0.3636               & 0.2516          & 0.4438       & 0.5781          & 0.2177       & \textbf{0.7622}    \\
{vertebral}   & 0.1318          & 0.0967           & 0.1489          & 0.1673          & 0.1072          & 0.0903          & 0.0918            & 0.1169               & 0.0999          & 0.1248       & 0.1022          & {\ul 0.2043} & \textbf{0.2088}    \\
{vowels}      & 0.3325          & 0.1619           & 0.3763          & 0.5163          & 0.0814          & 0.0341          & 0.0998            & {\ul 0.5506}         & 0.5272          & 0.1643       & 0.3595          & 0.3500       & \textbf{0.6643}    \\
{Waveform}    & 0.1236          & 0.0549           & 0.1311          & {\ul 0.2099}    & 0.0405          & 0.0559          & 0.0372            & 0.0489               & 0.1162          & 0.0638       & \textbf{0.4494} & 0.0506       & 0.1076             \\
{WBC}         & 0.5045          & {\ul 0.9388}     & 0.9198          & 0.9152          & 0.9038          & 0.9038          & 0.9040            & 0.5680               & 0.7120          & 0.1342       & 0.3530          & 0.0769       & \textbf{0.9642}    \\
{WDBC}        & 0.9573          & 0.6515           & 0.9472          & {\ul 0.9657}    & 0.5053          & 0.8037          & 0.5343            & 0.2650               & 0.3963          & 0.0527       & 0.9558          & 0.5259       & \textbf{0.9711}    \\
{Wilt}        & {\ul 0.1130}    & 0.0448           & 0.1115          & \textbf{0.1231} & 0.0417          & 0.0370          & 0.0373            & 0.0559               & 0.0498          & 0.0389       & 0.0533          & 0.0518       & 0.0839             \\
{wine}        & \textbf{0.9909} & 0.3014           & 0.9829          & 0.9851          & 0.1907          & 0.3612          & 0.2646            & 0.1579               & 0.0655          & 0.0869       & 0.4137          & 0.0843       & 0.9854             \\
{WPBC}        & 0.2316          & 0.2334           & 0.2316          & 0.2328          & 0.2178          & 0.2378          & 0.2192            & 0.2166               & 0.2251          & 0.2190       & \textbf{0.2603} & 0.2048       & {\ul 0.2479}       \\ \hline
\textbf{Average} &0.2765 & 0.3941 & 0.4336 & {\ul0.4414}  & 0.3462 & 0.3394 & 0.3344 & 0.2990 & 0.2723 & 0.3111 & 0.3244 & 0.2471 & \textbf{0.5000}

\\ \hline
\end{tabular}%
}
\end{table}

\begin{table}[]
\caption{Results in term of AUC-PR standard deviation.}
\label{tab:prc-std}
\resizebox{\textwidth}{!}{%
\begin{tabular}{llllllllllllll}
\hline
\textbf{Data}                      & \textbf{LOF}    & \textbf{iForest} & \textbf{iNNE}   & \textbf{IDK}    & \textbf{ECOD}   & \textbf{COPOD}  & \textbf{D.SVDD} & \textbf{AE} & \textbf{LUNAR}  & \textbf{DIF} & \textbf{SLAD}   & \textbf{DTE} & \textbf{VEAD} \\ \hline
ALOI                               & 0.0000       & 0.0003           & 0.0003        & 0.0013       & 0.0000        & 0.0000         & 0.0313            & 0.0000               & 0.0369         & 0.0029       & 0.0000        & 0.0007       & 0.0011         \\
annthyroid                         & 0.0000       & 0.0353           & 0.0063        & 0.0062       & 0.0000        & 0.0000         & 0.0120            & 0.0000               & 0.0158         & 0.0117       & 0.0000        & 0.0027       & 0.0058         \\
backdoor                           & 0.0000       & 0.0053           & 0.0068        & 0.0311       & 0.0000        & 0.0000         & 0.0547            & 0.0000               & 0.0053         & 0.0354       & 0.0000        & 0.0984       & 0.0187         \\
breastw                            & 0.0000       & 0.0017           & 0.0086        & 0.0004       & 0.0000        & 0.0000         & 0.0646            & 0.0000               & 0.0165         & 0.0239       & 0.0000        & 0.0306       & 0.0003         \\
campaign                           & 0.0000       & 0.0048           & 0.0048        & 0.0053       & 0.0000        & 0.0000         & 0.0413            & 0.0000               & 0.0178         & 0.0226       & 0.0000        & 0.0121       & 0.0082         \\
cardio                             & 0.0000       & 0.0471           & 0.0046        & 0.0216       & 0.0000        & 0.0000         & 0.0868            & 0.0000               & 0.0410         & 0.0299       & 0.0000        & 0.0224       & 0.0246         \\
Cardio2                            & 0.0000       & 0.0231           & 0.0109        & 0.0160       & 0.0000        & 0.0000         & 0.0376            & 0.0000               & 0.0153         & 0.0122       & 0.0000        & 0.0371       & 0.0170         \\
celeba                             & 0.0000       & 0.0056           & 0.0105        & 0.0061       & 0.0000        & 0.0000         & 0.0274            & 0.0000               & 0.0003         & 0.0037       & 0.0000        & 0.0111       & 0.0049         \\
census                             & 0.0000       & 0.0052           & 0.0056        & 0.0022       & 0.0000        & 0.0000         & 0.0043            & 0.0000               & 0.0012         & 0.0018       & 0.0000        & 0.0044       & 0.0013         \\
donors                             & 0.0000       & 0.0241           & 0.0185        & 0.0413       & 0.0000        & 0.0000         & 0.0565            & 0.0000               & 0.0002         & 0.0035       & 0.0000        & 0.0274       & 0.0023         \\
fault                              & 0.0000       & 0.0221           & 0.0083        & 0.0125       & 0.0000        & 0.0000         & 0.0545            & 0.0000               & 0.0083         & 0.0101       & 0.0000        & 0.0224       & 0.0040         \\
fraud                              & 0.0000       & 0.0769           & 0.0313        & 0.0208       & 0.0000        & 0.0000         & 0.0587            & 0.0000               & 0.0037         & 0.0244       & 0.0000        & 0.0185       & 0.0104         \\
glass                              & 0.0000       & 0.0137           & 0.0285        & 0.0187       & 0.0000        & 0.0000         & 0.0445            & 0.0000               & 0.0415         & 0.0248       & 0.0000        & 0.0107       & 0.0430         \\
Hepatitis                          & 0.0000       & 0.0306           & 0.0205        & 0.0170       & 0.0000        & 0.0000         & 0.0460            & 0.0000               & 0.0205         & 0.0602       & 0.0000        & 0.0176       & 0.0114         \\
http                               & 0.0000       & 0.0689           & 0.2367        & 0.1073       & 0.0000        & 0.0000         & 0.1265            & 0.0000               & 0.0000         & 0.0006       & 0.0000        & 0.2199       & 0.0000         \\
InternetADs                        & 0.0000       & 0.0445           & 0.0160        & 0.0080       & 0.0000        & 0.0000         & 0.0302            & 0.0000               & 0.0051         & 0.0119       & 0.0000        & 0.0458       & 0.0124         \\
Ionosphere                         & 0.0000       & 0.0032           & 0.0012        & 0.0098       & 0.0000        & 0.0000         & 0.0409            & 0.0000               & 0.0133         & 0.0066       & 0.0000        & 0.0117       & 0.0022         \\
landsat                            & 0.0000       & 0.0032           & 0.0065        & 0.0111       & 0.0000        & 0.0000         & 0.0609            & 0.0000               & 0.0026         & 0.0019       & 0.0000        & 0.0186       & 0.0041         \\
letter                             & 0.0000       & 0.0041           & 0.0093        & 0.0243       & 0.0000        & 0.0000         & 0.0061            & 0.0000               & 0.0220         & 0.0072       & 0.0000        & 0.0387       & 0.0290         \\
Lymphography                       & 0.0000       & 0.0270           & 0.0517        & 0.0390       & 0.0000        & 0.0000         & 0.1338            & 0.0000               & 0.0903         & 0.1444       & 0.0000        & 0.0934       & 0.0075         \\
magic.gamma & 0.0000       & 0.0046           & 0.0115        & 0.0105       & 0.0000        & 0.0000         & 0.0198            & 0.0000               & 0.0044         & 0.0070       & 0.0000        & 0.0090       & 0.0008         \\
mammography & 0.0000       & 0.0636           & 0.0120        & 0.0220       & 0.0000        & 0.0000         & 0.0679            & 0.0000               & 0.0115         & 0.0120       & 0.0000        & 0.0245       & 0.0109         \\
mnist       & 0.0000       & 0.0225           & 0.0357        & 0.0161       & 0.0000        & 0.0000         & 0.0708            & 0.0000               & 0.0305         & 0.0246       & 0.0000        & 0.0535       & 0.0075         \\
musk        & 0.0000       & 0.0122           & 0.0000        & 0.0000       & 0.0000        & 0.0000         & 0.3245            & 0.0000               & 0.0246         & 0.0727       & 0.0000        & 0.0385       & 0.0000         \\
optdigits   & 0.0000       & 0.0091           & 0.0137        & 0.0307       & 0.0000        & 0.0000         & 0.0067            & 0.0000               & 0.0031         & 0.0041       & 0.0000        & 0.0000       & 0.0085         \\
PageBlocks  & 0.0000       & 0.0106           & 0.0064        & 0.0245       & 0.0000        & 0.0000         & 0.0483            & 0.0000               & 0.0072         & 0.0216       & 0.0000        & 0.0799       & 0.0112         \\
pendigits   & 0.0000       & 0.0434           & 0.0145        & 0.0554       & 0.0000        & 0.0000         & 0.0757            & 0.0000               & 0.0081         & 0.0324       & 0.0000        & 0.0064       & 0.0417         \\
Pima        & 0.0000       & 0.0213           & 0.0025        & 0.0149       & 0.0000        & 0.0000         & 0.0314            & 0.0000               & 0.0155         & 0.0054       & 0.0000        & 0.0693       & 0.0041         \\
satellite   & 0.0000       & 0.0072           & 0.0162        & 0.0115       & 0.0000        & 0.0000         & 0.0600            & 0.0000               & 0.0027         & 0.0105       & 0.0000        & 0.0669       & 0.0095         \\
satimage-2  & 0.0000       & 0.0043           & 0.0039        & 0.0716       & 0.0000        & 0.0000         & 0.2241            & 0.0000               & 0.0354         & 0.0546       & 0.0000        & 0.0015       & 0.0010         \\
shuttle     & 0.0000       & 0.0042           & 0.0078        & 0.0065       & 0.0000        & 0.0000         & 0.0083            & 0.0000               & 0.0020         & 0.0530       & 0.0000        & 0.0000       & 0.0137         \\
skin        & 0.0000       & 0.0051           & 0.0292        & 0.0277       & 0.0000        & 0.0000         & 0.0187            & 0.0000               & 0.0005         & 0.0044       & 0.0000        & 0.0137       & 0.0119         \\
smtp        & 0.0000       & 0.0742           & 0.0431        & 0.0737       & 0.0000        & 0.0000         & 0.1021            & 0.0000               & 0.0479         & 0.0300       & 0.0000        & 0.0000       & 0.0085         \\
SpamBase    & 0.0000       & 0.0330           & 0.0088        & 0.0034       & 0.0000        & 0.0000         & 0.0217            & 0.0000               & 0.0105         & 0.0119       & 0.0000        & 0.1136       & 0.0025         \\
speech      & 0.0000       & 0.0011           & 0.0186        & 0.0108       & 0.0000        & 0.0000         & 0.0032            & 0.0000               & 0.0024         & 0.0028       & 0.0000        & 0.0050       & 0.0196         \\
Stamps      & 0.0000       & 0.0073           & 0.0218        & 0.0118       & 0.0000        & 0.0000         & 0.0812            & 0.0000               & 0.0269         & 0.0210       & 0.0000        & 0.0156       & 0.0181         \\
thyroid     & 0.0000       & 0.1061           & 0.0416        & 0.0246       & 0.0000        & 0.0000         & 0.0317            & 0.0000               & 0.0288         & 0.0245       & 0.0000        & 0.0173       & 0.0049         \\
vertebral   & 0.0000       & 0.0058           & 0.0000        & 0.0174       & 0.0000        & 0.0000         & 0.0059            & 0.0000               & 0.0038         & 0.0059       & 0.0000        & 0.0588       & 0.0095         \\
vowels      & 0.0000       & 0.0526           & 0.0524        & 0.0372       & 0.0000        & 0.0000         & 0.0698            & 0.0000               & 0.0769         & 0.0244       & 0.0000        & 0.0513       & 0.0333         \\
Waveform    & 0.0000       & 0.0039           & 0.0215        & 0.0213       & 0.0000        & 0.0000         & 0.0083            & 0.0000               & 0.0053         & 0.0053       & 0.0000        & 0.0050       & 0.0078         \\
WBC         & 0.0000       & 0.0063           & 0.0390        & 0.0159       & 0.0000        & 0.0000         & 0.0321            & 0.0000               & 0.1193         & 0.0173       & 0.0000        & 0.0125       & 0.0021         \\
WDBC        & 0.0000       & 0.0366           & 0.0150        & 0.1376       & 0.0000        & 0.0000         & 0.0677            & 0.0000               & 0.0677         & 0.0055       & 0.0000        & 0.4415       & 0.0099         \\
Wilt        & 0.0000       & 0.0020           & 0.0036        & 0.0059       & 0.0000        & 0.0000         & 0.0020            & 0.0000               & 0.0007         & 0.0014       & 0.0000        & 0.0182       & 0.0008         \\
wine        & 0.0000       & 0.0517           & 0.0040        & 0.0048       & 0.0000        & 0.0000         & 0.1540            & 0.0000               & 0.0054         & 0.0120       & 0.0000        & 0.0333       & 0.0046         \\
WPBC        & 0.0000       & 0.0231           & 0.0043        & 0.0044       & 0.0000        & 0.0000         & 0.0081            & 0.0000               & 0.0086         & 0.0053       & 0.0000        & 0.0086       & 0.0054      
\\ \hline
\end{tabular}%
}
\end{table}


\end{document}